\documentclass{article}
\usepackage{jointq_macros}
\title{\LARGE JointQ: Joint Scale-and-Integer Quantization \\ for Large Language Models}
\author{%
   Keiji Kimura\thanks{%
   Fujitsu Limited, Kanagawa, Japan, \@
   E-mail: kimura-keiji@fujitsu.com}\\
   Yuma Ichikawa\thanks{%
   Fujitsu Limited, Kanagawa, Japan, \@
   E-mail: ichikawa.yuma@fujitsu.com
   }
}
\date{}
\begin{document}
\maketitle
\begin{abstract}
\noindent
    GPTQ is the de facto standard solver for layer-wise post-training quantization (PTQ) of large language models, yet the solver itself has received less attention than weight preprocessing and cross-layer objectives.
    We target a bottleneck of GPTQ: its greedy procedure optimizes continuous group scales separately from the assignment of discrete integers.
    Evaluating each integer candidate with its closed-form optimal scale has been used for per-channel quantization; JointQ extends it to group-wise quantization, where a row's scales are coupled through the Hessian, and applies it to refine the GPTQ solution. Each local-search move changes one integer and recomputes its group's optimal scale in closed form, accounting for the other groups through the full Hessian; all scales are periodically re-solved jointly. The search starts from a row-wise selection among scale-refitted candidates including the GPTQ solution, so unregularized JointQ never has a larger layer-wise calibration error than GPTQ. JointQ preserves the GPTQ output format and is thus a drop-in replacement.
    Across four models and four settings, JointQ improves six-task average zero-shot accuracy over vanilla GPTQ in 14 of 16 comparisons and perplexity in 12 of 16. The mean accuracy gain increases from $0.0044$ in W4G128 to $0.1559$ in W3 per-channel, and JointQ also improves average accuracy over GPTQ with MSE-based scale search and activation ordering in all eight per-channel comparisons, though rarely in G128.
    An ablation shows that JointQ's diagonal Tikhonov regularization mitigates severe W3 per-channel perplexity degradation, so the largest gains are not due to the local search alone.
    Overall, JointQ's primary empirical benefit is downstream accuracy preservation in challenging low-bit and large-group regimes.
    Code is available in the OneComp library at \url{https://github.com/FujitsuResearch/OneCompression}.
\end{abstract}
\section{Introduction}
The rapid scaling of transformer-based language models has made efficient deployment a central challenge. Modern models now routinely contain billions to hundreds of billions of parameters, placing substantial pressure on memory capacity, memory bandwidth, and inference cost~\citep{vaswani2017attention,touvron2023llama2,dubey2024llama3,yang2025qwen3}. 
Post-training quantization (PTQ) is therefore a key technique for practical inference: it compresses pretrained weights from 16-bit precision to 4, 3, or even fewer bits without retraining, thereby reducing both model size and serving cost~\citep{gholami2022survey,nagel2021white}. 
Among PTQ methods, \emph{layer-wise} approaches are particularly attractive for large language models because each linear layer can be quantized independently using only a small calibration set, while the resulting model remains compatible with efficient low-bit inference kernels.

In this layer-wise setting, GPTQ~\citep{frantar2023gptq}, built on Optimal Brain Compression~\citep{frantar2022obc}, has become the de facto standard solver due to its strong speed--quality trade-off: it quantizes weight columns greedily using second-order activation statistics while compensating for the induced error in the remaining columns. It is widely used both as a standalone quantizer and as the final projection step in more elaborate PTQ pipelines~\citep{arai2025qep,li2025gptaq,lin2025loaq,ichikawa2025lpcd}. Recent progress in LLM quantization has largely advanced along two complementary directions. The first improves the geometry of weights and activations before quantization, for example, through rotation-based preprocessing methods~\citep{ashkboos2024quarot, liu2024spinquant, tseng2024qtip, tseng2024quipsharp, chee2023quip, lin2024duquant, sun2024flatquant} that mitigate outliers and make weight distributions more quantization-friendly. The second modifies the quantization objective to better account for calibration mismatch and error propagation, for example, through asymmetric calibration, output-aware formulations, or submodule-level correction~\citep{arai2025qep,li2025gptaq,lin2025loaq,ichikawa2025lpcd}. Although these advances have substantially improved PTQ performance, they mainly refine the context in which quantization is performed while leaving the core projector---the optimization step that maps a real-valued layer back to a low-bit group-wise quantized representation in the standard GPTQ format---largely unchanged.

We argue that this bottleneck becomes especially pronounced in challenging quantization regimes. In group-wise scalar quantization, the final quantized weights are governed by two tightly coupled components: continuous group scales and discrete integer assignments. The scales determine the spacing of the quantization grid, while the integers select the grid points. Standard GPTQ-style quantization optimizes these two components separately: it first sets the scales using a heuristic rule and then greedily assigns integers under the resulting fixed grid. Although this decoupling is often sufficient when the grid is dense or groups are small, it becomes increasingly restrictive in more challenging settings, such as low-bit quantization or large-group schemes like per-channel quantization, where the choice of scale significantly impacts reconstruction quality.

Joint treatments of the two components have so far been formulated mainly for a single scale per layer or per output channel. COMQ~\citep{zhang2025comq} alternates between integer updates under a fixed scale and closed-form scale updates. Beacon~\citep{zhang2025beacon} eliminates the scale analytically, obtaining a reduced objective that evaluates every integer candidate with its optimal scale, and searches over the integers under a cosine criterion derived from it. Neither addresses group-wise quantization, in which the scales of a row are coupled through the off-diagonal blocks of the Hessian, and both construct their integers independently of GPTQ. PiSO~\citep{amboage2026piso} computes optimal scales for round-to-nearest assignment and interleaves them with GPTQ, but the integers are then still assigned under fixed scales.

To address these gaps, we propose \textbf{JointQ}, a drop-in replacement for GPTQ that jointly optimizes scales and integer assignments while preserving the exact deployment format. JointQ rests on two design choices. First, its local search extends scale-eliminated integer search to the group-wise setting. Each candidate move modifies a single integer assignment and recomputes the optimal scale of the affected group in closed form, while accounting for the other groups through the full Hessian, so that the cross-group coupling enters the score of every move; after each pass over the groups, the search also re-solves all group scales of selected rows jointly with the integers fixed. Second, JointQ refines the GPTQ solution rather than replacing it. A GPTQ solution and a clipping-based solution are each given jointly refitted group scales, the candidate with the lower layer-local objective is selected independently for each output row, and the local search starts from this selection. Because the GPTQ solution is in the candidate pool, unregularized JointQ never has a larger layer-wise calibration error than GPTQ. In the per-channel case, the move score coincides with Beacon's reduced objective, so the distinction from such methods there lies in the second choice and in the regularization of Section~\ref{sec:regularization}. The resulting representation remains fully compatible with existing inference kernels and GPTQ-style checkpoints.

Empirically, across four models and four quantization configurations, JointQ improves six-task average zero-shot accuracy over vanilla GPTQ in 14 of 16 comparisons and perplexity in 12 of 16. The benefit grows with the difficulty of the quantization regime: averaged over the models, the accuracy gain increases from $0.0044$ in W4G128 to $0.1559$ in W3 per-channel. Against GPTQ with MSE-based scale search and activation ordering, JointQ wins all eight per-channel accuracy comparisons, with the clearest gains appearing in W3 per-channel. In G128, by contrast, the value of JointQ when its GPTQ candidate uses the same enhanced heuristics is model dependent; under the actorder-free evaluation condition, JointQ remains stronger than both vanilla and MSE-only GPTQ in most comparisons. A staged ablation separates the unregularized JointQ pipeline, consisting of the warm start and the local search, from the full stabilization policy: by construction, the unregularized pipeline never increases layer-wise calibration output error relative to GPTQ, and it reduces this error in every model, setting, and layer type shown, but the magnitude of this reduction does not predict downstream improvement. These findings are consistent with the mechanism exposed by our separation result, but the theory does not guarantee downstream-metric improvement. Perplexity also remains model dependent. We do not compare with COMQ, Beacon, or PiSO, so the experiments do not isolate the source of the per-channel gains, which are the largest (Section~\ref{sec:conclusion}). We therefore identify downstream accuracy preservation in challenging low-bit and large-group regimes, rather than uniform dominance on every metric, as the principal empirical benefit. The gains require no change to the standard GPTQ output format.
\section{Background}
\label{sec:background}

This section introduces the mathematical framework underlying both GPTQ and JointQ. We begin by defining the notation, presenting the layer-wise PTQ objective, describing the group-wise quantization format, and reviewing the OBQ and GPTQ algorithms that form the basis of JointQ.

\paragraph{Notation.}
We adopt row-vector notation, following the convention of~\citet{li2025gptaq}. Let $W \in \mab{R}^{m \times p}$ denote the weight matrix of a linear layer, stored row-wise by output channel, and let $X \in \mab{R}^{p \times n}$ denote the input activation matrix, where $m$ is the number of output channels, $p$ is the number of input neurons, and $n$ is the number of calibration tokens. The layer output is given by $Y = WX \in \mab{R}^{m \times n}$.
For a single output channel, we use $w \in \mab{R}^{1 \times p}$ to denote a row of $W$ and $\widehat{w} \in \mab{R}^{1 \times p}$ for its quantized counterpart. We write $e_j \in \mab{R}^{1 \times p}$ for the $j$-th standard basis row vector, such that $w e_j^\top = w_j$ extracts the $j$-th coordinate. Throughout the theoretical analysis, we define the regularized Hessian as $\widetilde{H} \coloneqq H + n\lambda R$, where $\lambda \geq 0$ is a regularization parameter and $R \succeq 0$ is the diagonal regularization matrix defined in Section~\ref{sec:regularization}. Results that require strict positive definiteness state $\widetilde{H} \succ 0$ explicitly.

\subsection{Layer-wise PTQ objective}
\label{subsec:layer-wise-ptq}
The goal of layer-wise PTQ is to find a quantized weight matrix $\widehat{W}$ that minimizes the reconstruction error of the layer output for the calibration data.
Formally, the problem is
\begin{equation}
  \min_{\widehat{W} \in \mac{Q}} \lVert WX - \widehat{W} X \rVert_F^2,
  \label{eq:layerwise_ptq}
\end{equation}
where $\mac{Q}$ denotes the set of representable quantized matrices determined by bit-width, grouping scheme, and quantization format.
Introducing the Hessian matrix $H \delequal XX^\top \in \mab{R}^{p \times p}$, the objective can be expressed as
\begin{equation}
  \lVert (W-\widehat{W})X \rVert_F^2
  = \tr\ab((W-\widehat{W}) H (W-\widehat{W})^\top).
  \label{eq:trace_form}
\end{equation}
The Hessian $H$ encodes second-order information about the input activations: its diagonal entry $H_{jj} = \sum_{i=1}^n X_{ji}^2$ measures the sensitivity of the layer output to perturbations of the $j$-th weight column, while the off-diagonal entry $H_{jk}$ captures the correlation between columns $j$ and $k$.
Columns with large $H_{jj}$ are important because quantization errors in these columns affect the output.

A crucial structural property is that the trace in Eq.~\eqref{eq:trace_form} decomposes row-wise.
Writing $w_i, \widehat{w}_i \in \mab{R}^{1 \times p}$ for the $i$-th rows of $W$ and $\widehat{W}$,
\begin{equation}
  \lVert (W-\widehat{W})X \rVert_F^2
  = \sum_{i=1}^m (w_i - \widehat{w}_i) H (w_i - \widehat{w}_i)^\top.
  \label{eq:rowwise_obj}
\end{equation}
This separability is fundamental: once $H$ is computed from the calibration data, the layer-wise PTQ problem reduces to $m$ independent row-wise quadratic programs, one for each output channel.
We denote the per-row objective by
\begin{equation}
  \phi(w, \widehat{w}) \delequal (w - \widehat{w}) H (w - \widehat{w})^\top \in \mab{R}.
  \label{eq:row_obj}
\end{equation}
Both GPTQ and JointQ exploit this row-wise separability, differing only in their methods for solving each row problem.

\subsection{Group-wise Uniform Scalar Quantization}
\label{sec:group_format}

The feasible set $\mac{Q}$ in Eq.~\eqref{eq:layerwise_ptq} is defined by the quantization format. In the standard group-wise scalar quantization scheme used by GPTQ, each row of $p$ weights is divided into $T$ contiguous groups $G_0, \dots, G_{T-1}$ of size $d$ where
\begin{equation}
    G_t = \{td, td{+}1, \dots, td{+}d{-}1\}, \qquad t = 0, \dots, T{-}1.
\end{equation}
Common choices include $d = 128$ and $d = p$; the latter refers to per-channel quantization. When activation ordering is enabled (Appendix~\ref{app:gptqstar}), the columns are first permuted and the groups are contiguous in the permuted order; all results below apply verbatim in the permuted coordinates.
For bit-width $b \ge 2$, let $L \coloneqq 2^{b-1}$. We use the fixed-zero-point format of the GPTQ reference implementation, in which each centered integer satisfies
\begin{equation}
  q_j \in \mac{C}_b \coloneqq \{-L, -L+1, \ldots, L-1\},
  \label{eq:codebook}
\end{equation}
and each group carries one scale, so that $\widehat{w}_{G_t} = s_t q_t$ with $q_t \in \mac{C}_b^d$. Following common usage, we call this format \emph{symmetric}; the term refers to the zero point being fixed at the midpoint of the raw range, not to $\mac{C}_b$ being symmetric about zero. Indeed, $\mac{C}_b$ has $L$ negative levels and $L-1$ positive levels. In storage, the centered integers are offset as $q_j^{\mathrm{raw}} = q_j + L \in \{0, \ldots, 2^b - 1\}$, so all $2^b$ raw levels are used. The asymmetric format, in which stored raw integers $q_j^{\mathrm{raw}}, z_t \in \{0, \ldots, 2^b - 1\}$ define centered integers $\bar{q}_j = q_j^{\mathrm{raw}} - z_t$, is deferred to Appendix~\ref{app:asymmetric}.

In standard practice, the group scale is selected by the min--max rule
\begin{equation}
  s_t = \frac{2 \max_{j \in G_t} |w_j|}{2^b - 1}
      = \frac{\max_{j \in G_t} |w_j|}{L - \tfrac{1}{2}},
  \label{eq:scale_minmax}
\end{equation}
followed by $q_j = \mathrm{clamp}(\lfloor w_j / s_t \rceil, -L, L-1)$, where $\lfloor \cdot \rceil$ rounds half to even. Under this rule, the entry of largest magnitude is mapped to $\pm(L - \tfrac{1}{2})$, exactly halfway between two levels, so it is never reproduced exactly. After rounding and clipping, a positive extreme is reconstructed as $\max_j |w_j| \cdot (L-1)/(L-\tfrac{1}{2})$ and, under round-half-to-even, a negative extreme as $-\max_j |w_j| \cdot L/(L-\tfrac{1}{2})$; the error is $s_t/2$ in either case. The rule thus fixes the grid from a single extreme value and can use the available levels inefficiently, particularly when outliers are present.

A central observation in this paper is that the quantized row $\widehat{w}$ is determined by two distinct but tightly coupled sets of variables: the continuous group scales $\{s_t\}_{t=0}^{T-1}$ and the discrete integers $\{q_j\}_{j=0}^{p-1}$. Standard GPTQ treats these components separately, whereas JointQ optimizes them jointly.

\subsection{OBQ and GPTQ}
\label{sec:gptq_background}
We review how GPTQ addresses the row-wise objective in Eq.~\eqref{eq:row_obj}. GPTQ~\citep{frantar2023gptq} builds on the Optimal Brain Quantization (OBQ) framework~\citep{frantar2022obc}, which extends the classical Optimal Brain Surgeon method~\citep{hassibi1993optimal} for quantization. 

We assume that the solver uses a positive-definite Hessian surrogate. In practice, GPTQ regularizes the Hessian as $H \leftarrow H + \epsilon \cdot \mathrm{mean}(\mathrm{diag}(H)) I_p$ before inversion. OBQ then quantizes one coordinate at a time. When quantizing the $l$-th coordinate of a row $w$, it seeks a perturbation $\Delta w = \widehat{w} - w \in \mab{R}^{1 \times p}$ that minimizes the row-wise reconstruction error, subject to the constraint that coordinate $l$ is fixed at its quantized value, namely $\Delta w e_l^\top = \widehat{w}_l - w_l$. The resulting Lagrangian optimality conditions yield the closed-form update
\begin{equation}
  \Delta w = \frac{(\widehat{w}_l - w_l)}{[H^{-1}]_{ll}} \cdot [H^{-1}]_{l,:},~~~
  \mac{E}_l = \frac{(\widehat{w}_l - w_l)^2}{[H^{-1}]_{ll}}.
  \label{eq:obq_update}
\end{equation}
The first expression indicates that the quantization error introduced at coordinate $l$ is redistributed across the remaining coordinates according to the $l$-th row of the inverse Hessian. The second gives the associated per-coordinate loss, which is larger when $[H^{-1}]_{ll}$ is small, indicating that coordinate $l$ is more sensitive. After quantizing coordinate $l$, the inverse Hessian over the remaining coordinates $\bar{l} = \{0,\dots,p{-}1\} \setminus \{l\}$ is updated via the Schur complement:
\begin{equation}
  [H^{-1}]_{\bar{l},\bar{l}} \leftarrow [H^{-1}]_{\bar{l},\bar{l}} - \frac{[H^{-1}]_{\bar{l},l} [H^{-1}]_{l,\bar{l}}}{[H^{-1}]_{ll}}.
  \label{eq:hessian_update}
\end{equation}
Although OBQ provides closed-form optimal updates, it is too expensive for billion-parameter models because its greedy ordering based on $\mac{E}_l$ hinders efficient parallelization across output channels. GPTQ~\citep{frantar2023gptq} makes this framework practical through three key modifications. First, instead of following the greedy ordering induced by $\mac{E}_l$, it processes coordinates in a fixed column order $l = 0, 1, \dots, p-1$, enabling parallel quantization across all $m$ output channels. Second, it accumulates updates in blocks of size $B$ and applies them lazily, which raises the compute-to-memory ratio and improves hardware efficiency. Third, instead of repeatedly applying the Schur-complement update above, which accumulates numerical errors at billion-parameter scale, it precomputes the required rows of the updated inverse Hessians from a single Cholesky factorization of $H^{-1}$, improving numerical stability.

Despite its effectiveness, GPTQ does not jointly optimize the continuous group scales $\{s_t\}$ and the discrete integer assignments $\{q_j\}$. Each group scale is set by Eq.~\eqref{eq:scale_minmax} before any integer of the group is assigned, and once the greedy column-wise pass fixes the integer assignments, the scales are not revisited; Appendix~\ref{app:gptqstar} describes at which point of the pass each scale is computed. This decoupling is often harmless when the quantization grid is sufficiently dense, such as at higher bit-widths or with small groups. However, it becomes much more restrictive in low-bit or per-channel settings, where each group contains only a few available grid points and scale selection significantly affects reconstruction quality. The next section introduces JointQ, which directly addresses this limitation.
\section{JointQ}
\label{sec:jointq}
This section introduces JointQ. We begin by formulating the coupled row-wise optimization problem and showing that the continuous scale variables can be eliminated analytically, yielding a reduced discrete reference objective. We then present the group-conditional surrogate used by the practical algorithm, derive the corresponding closed-form best-response scale, and decompose the resulting conditional error into a direction gap and a scale gap. Next, we provide an illustrative separation example for a single-group instance, showing that frozen-scale quantization can be fundamentally suboptimal in low-bit regimes. Finally, we describe the local search procedure, regularization strategy, and implementation details.

\subsection{Problem formulation}
\label{sec:formulation}
JointQ targets the same quantization format as GPTQ, but differs in one essential respect: it treats group scales and integer assignments as coupled optimization variables rather than optimizing them separately. By Eq.~\eqref{eq:rowwise_obj}, the layer-wise objective decomposes across output channels, so analyzing a single row $w \in \mab{R}^{1 \times p}$ is sufficient. For notational convenience, we assume that $d$ divides $p$; otherwise, the final group may be smaller, and all results extend without modification. Each row is partitioned into $T = p/d$ groups, with associated scales $\B{s} = (s_0, \dots, s_{T-1}) \in \mab{R}^T$ and integer assignments $q_j \in \mac{C}_b$ for all $j$. In the theoretical development, we use the fixed-zero-point codebook $\mac{C}_b = \{-L, \ldots, L-1\}$ of Section~\ref{sec:group_format}. None of the results in this section requires the codebook to be symmetric about zero, except Proposition~\ref{prop:separation}, which is an illustrative example stated for the symmetric subcodebook $\mac{C}_b^{\mathrm{sym}} \coloneqq \{-(L-1), \ldots, L-1\} \subset \mac{C}_b$. The extension to asymmetric quantization is presented in Appendix~\ref{app:asymmetric}, and the zero-point used in the implementation is described in Section~\ref{sec:implementation}.

For each group $G_t$, let $a_t \in \mab{R}^{1 \times p}$ denote the zero-padded vector with entries $q_j$ for $j \in G_t$ and $0$ elsewhere. The quantized row can be expressed as $\widehat{w} = \sum_{t=0}^{T-1} s_t a_t$, resulting in the row-wise coupled objective being
\begin{equation}
  \phi(\B{s}, \B{q}) = \ab(w - \textstyle\sum_{t=0}^{T-1} s_t a_t) \widetilde{H} \ab(w - \textstyle\sum_{t=0}^{T-1} s_t a_t)^\top.
  \label{eq:phi_def}
\end{equation}
Here, we use the regularized Hessian $\widetilde{H} = H + n\lambda R$ from the outset, where the diagonal matrix $R$ is defined in Section~\ref{sec:regularization}. We allow signed scales $s_t \in \mab{R}$. Because $\mac{C}_b$ is not symmetric about zero, this is not without loss of generality: a pair $(s_t, q_t)$ with $s_t < 0$ represents the same dequantized group as $(|s_t|, -q_t)$, where $-q_t \in \{-(L-1), \ldots, L\}^d$ need not lie in $\mac{C}_b^d$. Allowing $s_t < 0$ therefore lets each group choose on which side of zero the extra level lies. The best-response scale in Proposition~\ref{prop:best_scale} may be negative, and the implementation stores it as computed; the GPTQ-format dequantization $s_t(q_j^{\mathrm{raw}} - L)$ is well defined for either sign and requires no change to the storage format.

\subsection{Scale Elimination and Reduced Objective}
\label{sec:scale_elim}
The continuous scale variables in Eq.~\eqref{eq:phi_def} can be eliminated analytically, reducing the coupled discrete-continuous problem to a purely discrete optimization over integer assignments. For a single scale per output channel, this elimination is known~\citep{zhang2025beacon}; Theorem~\ref{thm:scale_elim} states it for $T$ group scales, which remain coupled through the off-diagonal blocks of $\widetilde{H}$. The elimination is more than a technical simplification; it demonstrates that the core difficulty of group-wise quantization lies in the discrete search itself, while the optimal scales are implicitly determined once the integers are fixed.

Formally, let $A(\B{q}) \in \mab{R}^{T \times p}$ denote the matrix whose $t$-th row is $a_t$, and define the matrix $B(\B{q}) \coloneqq A(\B{q}) \widetilde{H} A(\B{q})^\top \in \mab{R}^{T \times T}$.

\begin{theorem}[Global Scale Elimination]
\label{thm:scale_elim}
Assume $\widetilde{H} \succ 0$. Let $I(\B{q}) = \{t : a_t \neq 0\}$ denote the set of active groups. If $I(\B{q}) = \emptyset$, then $F(\B{q}) = w \widetilde{H} w^\top$, and every scale vector is optimal. If $I(\B{q}) \neq \emptyset$, restricting $A(\B{q})$ to its active rows yields a matrix $A_I(\B{q})$ such that $B_I(\B{q}) \coloneqq A_I \widetilde{H} A_I^\top$ is positive definite. In this case, the active components of every minimizer of Eq.~\eqref{eq:phi_def} are uniquely defined by
\begin{equation}
  \B{s}_I^\star(\B{q}) = w \widetilde{H} A_I(\B{q})^\top B_I(\B{q})^{-1},
  \label{eq:global_scale}
\end{equation}
while the inactive scales remain unconstrained; throughout the paper, we adopt the canonical choice $s_t^\star = 0$ for $t \notin I(\B{q})$. The resulting \emph{reduced objective} is
\begin{equation}
  F(\B{q}) \coloneqq \min_{\B{s}} \phi(\B{s}, \B{q})
  = w \widetilde{H} w^\top - w \widetilde{H} A_I(\B{q})^\top B_I(\B{q})^{-1} A_I(\B{q}) \widetilde{H} w^\top.
  \label{eq:reduced_obj}
\end{equation}
\end{theorem}
The proof can be found in Appendix~\ref{app:scale_elim_proof}.
The reduced objective $F(\B{q})$ in Eq.~\eqref{eq:reduced_obj} can be interpreted as the total energy minus the energy captured by the integer directions, that is, as a projection residual. The better the integer vectors $\{a_t\}$ span the weight vector $w$ under the $\widetilde{H}$-induced geometry, the smaller $F(\B{q})$ becomes. For $T > 1$, $B_I(\B{q})$ is in general not diagonal; its off-diagonal entries $a_t \widetilde{H} a_u^\top$ couple the groups. This also shows that $F(\B{q})$ provides a natural reference lower bound: for any scale vector $\B{s}$, we have $\phi(\B{s}, \B{q}) \ge F(\B{q})$, with equality if and only if the active scales are set to $\B{s}_I^\star(\B{q})$; the inactive scales remain arbitrary. JointQ uses Eq.~\eqref{eq:global_scale} in two places: in its warm start (Section~\ref{sec:warm_start}), where it refits the scales of each candidate solution before the candidates are ranked, and in the joint scale re-solve of the local search (Section~\ref{sec:local_search}). The integer moves of the local search, however, are not scored by $F$. Instead, they use the group-conditional surrogate introduced in Section~\ref{sec:best_response}. This perspective highlights a fundamental difference between JointQ and approaches that optimize scales and integer assignments in separate stages.

\paragraph{Positive Definiteness of $B_I(\B{q})$.}
Among the active rows $\{a_t : t \in I(\B{q})\}$, each has support from a distinct group $G_t$ and is nonzero by definition; therefore, any nonzero linear combination is also nonzero.
Since $\widetilde{H} \succ 0$, the $\widetilde{H}$-quadratic form evaluates to a strictly positive value, which results in $B_I(\B{q}) \succ 0$.
The detailed argument is found in Appendix~\ref{app:scale_elim_proof}.

\subsection{Best-response scale update}
\label{sec:best_response}
JointQ does not solve the entire $T \times T$ linear system in Eq.~\eqref{eq:global_scale} for every candidate move, as doing so would be prohibitively expensive when $T$ is large. Instead, it scores each move with a closed-form best-response rule that re-optimizes only the scale of the affected group while keeping all other groups fixed. The full system is solved only once per pass of the local search, and only for a subset of rows (Section~\ref{sec:local_search}).
Specifically, fix all groups except for group $t$, and define the residual
\begin{equation}
  r_t \delequal w - \sum_{u \neq t} s_u a_u \in \mab{R}^{1 \times p}.
  \label{eq:residual_def}
\end{equation}
Intuitively, $r_t$ indicates the portion of $w$ that remains unexplained by other groups. Under this conditioning, the objective becomes a scalar quadratic function of $s_t$:
\begin{equation}
  \phi(s_t) = r_t \widetilde{H} r_t^\top - 2 s_t(a_t \widetilde{H} r_t^\top) + s_t^2(a_t \widetilde{H} a_t^\top).
  \label{eq:one_group_quad}
\end{equation}

\begin{proposition}[Best-Response Scale]
\label{prop:best_scale}
If $a_t \widetilde{H} a_t^\top > 0$, then the unique minimizer of Eq.~\eqref{eq:one_group_quad} over $s_t \in \mab{R}$ is
\begin{equation}
  s_t^{\star} = \frac{a_t \widetilde{H} r_t^\top}{a_t \widetilde{H} a_t^\top},
  \qquad
  \phi(s_t^{\star}) = r_t \widetilde{H} r_t^\top - \frac{(a_t \widetilde{H} r_t^\top)^2}{a_t \widetilde{H} a_t^\top}.
  \label{eq:best_scale}
\end{equation}
If $a_t \widetilde{H} a_t^\top = 0$, we set $s_t^{\star} = 0$. The proof is included in Appendix~\ref{app:best_scale_proof}.
\end{proposition}

The best-response scale $s_t^\star$ can be viewed as a weighted regression coefficient; it projects the residual $r_t$ onto the integer direction $a_t$ using the $\widetilde{H}$-inner product. With $T = 1$ and $r_0 = w$, Eq.~\eqref{eq:best_scale} is the closed-form scale used by COMQ and Beacon~\citep{zhang2025comq,zhang2025beacon}; the residual $r_t$ extends it to a group whose neighbors are held fixed. The minimized value $\phi(s_t^\star)$ represents the residual after this projection. This expression clarifies the scale--integer coupling: changing the integer assignments alters the direction $a_t$, which in turn impacts the optimal scale. Standard GPTQ does not capture this dependence because it fixes the scale before optimizing the integers.

\paragraph{Group-Conditional Surrogate.}
The best-response update naturally leads to the quantity used by the practical algorithm. Fix all groups except $t$, and define the group-conditional surrogate
\begin{equation}
  \mac{L}_t(u) \delequal \min_{\sigma \in \mab{R}} \phi\ab((\B{s}_{-t}, \sigma), (\B{q}_{-t}, u)),
  \label{eq:Lt_def}
\end{equation}
where $\B{s}_{-t}$ and $\B{q}_{-t}$ denote the current scales and integer assignments for all groups except $t$, and $u \in \mac{C}_b^d$ is a candidate integer vector for group $t$. Let $\alpha_t \coloneqq r_t \widetilde{H} r_t^\top$, which is independent of $u$, and define
\begin{equation}
    \beta_t(u) \coloneqq a_t(u) \widetilde{H} r_t^\top,~~
    \gamma_t(u) \coloneqq a_t(u) \widetilde{H} a_t(u)^\top,
\end{equation}
where $a_t(u)$ is the zero-padded vector formed by placing $u$ at $G_t$. Then Proposition~\ref{prop:best_scale} gives
\begin{equation}
  \mac{L}_t(u) =
  \begin{cases}
    \alpha_t - \dfrac{\beta_t(u)^2}{\gamma_t(u)}, & \gamma_t(u) > 0, \\
    \alpha_t, & \gamma_t(u) = 0.
  \end{cases}
\end{equation}
When $\widetilde{H} \succ 0$, every nonzero $u$ satisfies $\gamma_t(u) > 0$, so the second case occurs only for $u = 0$. The explicit zero-curvature branch remains necessary when $\widetilde{H}$ is only positive semidefinite.

Algorithm~\ref{alg:inner} evaluates this surrogate $\mac{L}_t$ for every candidate move rather than the globally reduced objective $F$. This distinction is important: $F(\B{q}')$ jointly optimizes all $T$ scales, whereas $\mac{L}_t(u)$ re-optimizes only the scale of group $t$. As a result, for $\B{q}' = (\B{q}_{-t}, u)$, we always have $F(\B{q}') \le \mac{L}_t(u)$. For $T = 1$, the residual is $r_0 = w$ and the two coincide, $\mac{L}_0(u) = F(u)$, so in the per-channel case the moves are scored by the same reduced objective as in Beacon~\citep{zhang2025beacon}, with $\widetilde{H}$ in place of $H$. For $T > 1$, $\mac{L}_t$ also differs from a group-local objective that ignores the other groups: through $r_t$, the quantity $\beta_t(u)$ depends on the current reconstruction of every other group via the off-diagonal blocks of $\widetilde{H}$, so the cross-group coupling enters every move although only one scale is re-optimized.

\subsection{Error Decomposition}
\label{sec:error_decomp}
We next derive an exact decomposition of the single group conditional error for a fixed residual $r_t$, illustrating the advantage of joint scale--integer optimization over frozen-scale methods within a single group. The decomposition separates the error into two independent, non-negative terms, each with a clear geometric interpretation.
In this subsection, group-level vectors such as $x$, $q_t$, $v_t$, and $g_t$ are treated as column vectors in $\mab{R}^d$. Let $\widetilde{H}_t \coloneqq [\widetilde{H}]_{G_t, G_t} \in \mab{R}^{d \times d}$ denote the principal submatrix of $\widetilde{H}$ restricted to group $t$, and let $g_t \coloneqq [\widetilde{H} r_t^\top]_{G_t} \in \mab{R}^d$ denote the corresponding subvector. We define the group-level error function
\begin{equation}
    \Psi_t(x) \coloneqq r_t \widetilde{H} r_t^\top - 2 x^\top g_t + x^\top \widetilde{H}_t x,~~x \in \mab{R}^d.
\end{equation}

\begin{theorem}[Conditional Direction-Gap and Scale-Gap Decomposition]
\label{thm:decomp}
    Fix a residual $r_t$ as in Eq.~\eqref{eq:residual_def} and assume $\widetilde{H}_t \succ 0$. For any nonzero $q_t \in \mac{C}_b^d$ and any $s \in \mab{R}$, the excess error relative to the unrestricted continuous optimum admits the decomposition
    \begin{equation}
      \Psi_t(s q_t) - \Psi_t^{\mathrm{cont}}
      = \underbrace{\ab(\|v_t\|_{\widetilde{H}_t}^2 - \ab(\nicefrac{\langle q_t, v_t\rangle_{\widetilde{H}_t}^2}{\|q_t\|_{\widetilde{H}_t}^2}))}_{\text{direction gap}}
      + \underbrace{(q_t^\top \widetilde{H}_t q_t)\bigl(s - s_t^\star(q_t)\bigr)^2}_{\text{scale gap}},
      \label{eq:decomp}
    \end{equation}
    where $\Psi_t^{\mathrm{cont}} \coloneqq r_t \widetilde{H} r_t^\top - g_t^\top \widetilde{H}_t^{-1} g_t$ is the continuous minimum, $v_t \coloneqq \widetilde{H}_t^{-1} g_t$ is the unrestricted continuous optimizer, and $s_t^\star(q_t) \coloneqq (\nicefrac{q_t^\top g_t}{q_t^\top \widetilde{H}_t q_t})$
    is the best-response scale. We use the notation $\langle q_t, v_t\rangle_{\widetilde{H}_t} = q_t^\top \widetilde{H}_t v_t$ and $\|q_t\|_{\widetilde{H}_t}^2 = q_t^\top \widetilde{H}_t q_t$. If $v_t \neq 0$, the direction gap can be expressed as $\|v_t\|_{\widetilde{H}_t}^2 \sin^2 \theta_{\widetilde{H}_t}(q_t, v_t)$. If $v_t = 0$ or equivalently $g_t = 0$, then the unrestricted continuous optimizer is $x_t^\star = 0$, the continuous minimum is $\Psi_t^{\mathrm{cont}} = r_t \widetilde{H} r_t^\top$, and the direction gap vanishes. 
\end{theorem}
The proof is in Appendix~\ref{app:decomp_proof}. For a fixed residual $r_t$, Eq.~\eqref{eq:decomp} identifies two distinct sources of single-group conditional error. The direction gap represents the cost of integrality: even with the optimal scale, the integer direction $q_t$ cannot, in general, align perfectly with the continuous optimum $v_t$. Geometrically, this term is proportional to $\sin^2 \theta$, where $\theta$ is the angle between $q_t$ and $v_t$ under the Hessian-induced geometry. This error is inherent to the discrete codebook and cannot be removed by scale optimization alone. By contrast, the scale gap represents the cost of selecting a suboptimal scale. It takes the form $\gamma (s - s_t^\star)^2$, where $\gamma = q_t^\top \widetilde{H}_t q_t > 0$. JointQ eliminates this term for the updated group at each proposal by recomputing $s_t^\star$ in closed form, as Beacon does for a single scale per channel~\citep{zhang2025beacon}. GPTQ, CDQuant, and COMQ~\citep{nair2024cdquant,zhang2025comq} evaluate integer updates under a frozen scale and therefore incur both contributions at each such update. The following corollary explicitly states the advantage over frozen-scale evaluation.

\begin{corollary}[Dominance over Frozen-Scale Evaluation]
    \label{cor:frozen}
    For any candidate $q_t' \in \mac{C}_b^d$, let $s_t^\star(q_t')$ denote the best-response scale, following the convention $s_t^\star(0) = 0$. Then
    \begin{equation}
      \Psi_t(s_t q_t') - \Psi_t(s_t^\star(q_t') q_t')
      = (q_t'^\top \widetilde{H}_t q_t') \bigl(s_t - s_t^\star(q_t')\bigr)^2.
      \label{eq:frozen_dom}
    \end{equation}
    For $q_t' \neq 0$, this follows directly from Theorem~\ref{thm:decomp}, as both evaluations share the same direction gap. For $q_t' = 0$, both sides equal zero under the convention $s_t^\star(0) = 0$.
\end{corollary}

\subsection{Separation in Low-Bit Regimes}
\label{sec:separation}

Theorem~\ref{thm:decomp} demonstrates that joint optimization can qualitatively outperform the standard decoupled procedure. The following result makes this point concrete in a simple single-group setting. It constructs an explicit instance on the symmetric subcodebook $\mac{C}_b^{\mathrm{sym}}$ in which the ratio between the error of a min--max initialization and that of a nearby jointly optimized solution exceeds $(Q^2 + d - 1)/(Q^2 + 1)$, where $Q \coloneqq L - 1 = 2^{b-1} - 1$, and thus grows linearly with the group size $d$ and is of order $d/Q^2$. It also shows that the $\pm 1$ local search used by JointQ reaches the improved solution through a monotone sequence of strictly improving moves.

\begin{proposition}[Separation on the symmetric subcodebook]
\label{prop:separation}
Let $b \ge 2$ and $Q \coloneqq 2^{b-1} - 1$. Consider a single group ($T = 1$) of size $d \ge 2$ with $\widetilde{H} = I_d$ and the symmetric subcodebook $\mac{C}_b^{\mathrm{sym}} = \{-Q, \ldots, Q\}$. Let $w = (M, 1, \ldots, 1) \in \mab{R}^{1 \times d}$, where
\begin{equation*}
  2Q < M < Q + \sqrt{Q^2 + 1}.
\end{equation*}
For $q \in (\mac{C}_b^{\mathrm{sym}})^d$, define $E(q) \delequal \min_{s \in \mab{R}} \lVert w - sq\rVert_2^2$. Let $q^{(0)}$ be obtained by the symmetric min--max scale $s^{(0)} = \max_j |w_j| / Q$ followed by nearest-integer rounding and clipping to $[-Q, Q]$, and define $E_0 \delequal E(q^{(0)})$. Then $q^{(0)} = (Q, 0, \ldots, 0)$ and $E_0 = d - 1$. Moreover, there exists a sequence $q^{(1)}, \ldots, q^{(d-1)} \in (\mac{C}_b^{\mathrm{sym}})^d$ such that each $q^{(k+1)}$ differs from $q^{(k)}$ by a single $\pm 1$ update, the errors $E_k \delequal E(q^{(k)})$ decrease strictly, and
\begin{equation}
  E_{d-1} = \frac{(M - Q)^2 (d - 1)}{Q^2 + d - 1},
  \qquad
  \frac{E_0}{E_{d-1}} = \frac{Q^2 + d - 1}{(M - Q)^2}
  > \frac{Q^2 + d - 1}{Q^2 + 1}.
  \label{eq:separation}
\end{equation}
In particular, $E_0 / E_{d-1} = \Omega(d / Q^2)$. Finally, consider Algorithm~\ref{alg:inner} run on a single row ($m = 1$) with $g_{0,0} = w^\top$, initial integers $q^{(0)}$, any $\omega_{\mathrm{in}} \ge 0$, and $\varepsilon < \min_{0 \le k < d-1} (E_k - E_{k+1})$, for example $\varepsilon = 0$. Each of its first $d-1$ iterations raises one coordinate from $0$ to $1$, so its iterates coincide with $q^{(1)}, \ldots, q^{(d-1)}$ up to a permutation of the coordinates $1, \ldots, d-1$, which leaves every $E_k$ unchanged. This holds whether its feasibility check uses $\mac{C}_b^{\mathrm{sym}}$ or $\mac{C}_b$.
\end{proposition}
The proof, including the explicit monotone path, is provided in Appendix~\ref{app:separation_proof}.
Proposition~\ref{prop:separation} provides a mechanism consistent with the empirical trend in Section~\ref{sec:experiments}: the separation grows with the group size $d$ and is larger when the bit-width is small. For $b = 2$ ($Q = 1$), the admissible range is $2 < M < 1 + \sqrt{2}$, and the ratio exceeds $d/2$, which grows as $\Omega(d)$. The proposition is an idealized example. It uses the symmetric subcodebook and the symmetric min--max scale $\max_j |w_j|/Q$ rather than the GPTQ rule in Eq.~\eqref{eq:scale_minmax}. This scale is the one used by the Clip-Optimize candidate of JointQ's warm start (Section~\ref{sec:warm_start}); because $|w_j|/s^{(0)} \le Q$, clipping to $\mac{C}_b$ never binds, so $q^{(0)}$ is also the initial Clip-Optimize assignment, and with $T = 1$ and $\widetilde{H} = I_d$ the group-local refinement of that candidate is exactly the run of Algorithm~\ref{alg:inner} described in the proposition. Because $\mac{C}_b^{\mathrm{sym}} \subset \mac{C}_b$, every iterate in the constructed sequence is also a valid solution in the implemented format. The scope is deliberately limited: the result concerns a single group with $\widetilde{H} = I_d$ and min--max initialization, and does not guarantee monotonic gains for multi-group GPTQ, downstream accuracy, or the regularized full system. Within this setting, the group-conditional surrogate coincides with the reduced objective, demonstrating that a better jointly optimized solution exists and that JointQ's strictly improving $\pm 1$ updates reach it. Because $T = 1$, the proposition illustrates the gap between frozen min--max scaling and any scale-eliminated per-channel search, such as Beacon~\citep{zhang2025beacon}, rather than an advantage of JointQ over such searches.

\subsection{Local Search Procedure}
\label{sec:local_search}

Having established the underlying motivation, we describe the practical JointQ algorithm. Starting from any feasible GPTQ-format solution $(\B{s}^{(0)}, \B{q}^{(0)})$, in all our experiments the multi-candidate warm start of Section~\ref{sec:warm_start}, JointQ improves all $m$ rows of a layer simultaneously. We write $q_{i,t} \in \mac{C}_b^d$ and $s_{i,t} \in \mab{R}$ for the integers and the scale of group $t$ in row $i$, $r_{i,t}$ for the residual of Eq.~\eqref{eq:residual_def} of row $i$, and $\phi_i$ for its row objective in Eq.~\eqref{eq:phi_def}. Algorithm~\ref{alg:jointq} is a block-coordinate descent over the groups. One \emph{pass} visits the groups $t = 0, \dots, T-1$ in order and, at each visit, runs the integer search of Algorithm~\ref{alg:inner} on group $t$ while all other groups are fixed. After each pass, the scales of a subset of rows are re-solved jointly, and passes are repeated until few rows still improve.

\paragraph{Integer search within a group.}
During the visit to group $t$, the residual $r_{i,t}$ depends only on the other groups and is therefore fixed. So are $\alpha_t$ and the vector $g_{i,t} \coloneqq [\widetilde{H} r_{i,t}^\top]_{G_t} \in \mab{R}^d$ of Section~\ref{sec:error_decomp}; for all rows at once, the vectors $g_{i,t}^\top$ are the rows of $(W - \widehat{W})\widetilde{H}_{:,G_t} + \widehat{W}_{:,G_t}\widetilde{H}_t$, which takes a single matrix product. Suppressing the row index in $\alpha_t$, $\beta_t$, and $\gamma_t$, a candidate $u \in \mac{C}_b^d$ has $\beta_t(u) = u^\top g_{i,t}$ and $\gamma_t(u) = u^\top \widetilde{H}_t u$ in the notation of Section~\ref{sec:best_response}. We define the energy captured by $u$ as
\begin{equation}
  \kappa_t(u) \coloneqq
  \begin{cases}
    \beta_t(u)^2/\gamma_t(u), & \gamma_t(u) > \varepsilon, \\
    0, & \text{otherwise},
  \end{cases}
  \label{eq:captured}
\end{equation}
where $\varepsilon \ge 0$ is a numerical tolerance. Under the condition on $\varepsilon$ in Proposition~\ref{prop:termination}, $\mac{L}_t(u) = \alpha_t - \kappa_t(u)$, so minimizing the surrogate amounts to maximizing $\kappa_t$. Algorithm~\ref{alg:inner} keeps $\beta_t$, $\gamma_t$, and the cache $c_{i,t} \coloneqq \widetilde{H}_t u = [\widetilde{H} a_t(u)^\top]_{G_t} \in \mab{R}^d$ of the current $u$, so a move $u_j \leftarrow u_j + \delta$ with $\delta \in \{-1, +1\}$ is scored in $\mac{O}(1)$ time through
\begin{equation}
  \beta_t' = \beta_t + \delta [g_{i,t}]_j,
  \qquad
  \gamma_t' = \gamma_t + 2\delta [c_{i,t}]_j + [\widetilde{H}_t]_{jj}.
  \label{eq:incr}
\end{equation}
The search uses best improvement. In each iteration, it scores every feasible move of every row, takes for each row the best $+1$ move and the best $-1$ move, and applies the better of the two, choosing the $+1$ move on ties, if it increases $\kappa_t$ by more than $\varepsilon$. An applied move also updates the cache as $c_{i,t} \leftarrow c_{i,t} + \delta [\widetilde{H}_t]_{:,j}$ in $\mac{O}(d)$ time. Rows without such a move are left unchanged in that iteration. Writing $\Delta_i$ for the largest increase of $\kappa_t$ available to row $i$ in an iteration, the search stops once more than a fraction $\omega_{\mathrm{in}}$ of the rows have $\Delta_i < \varepsilon$, or after $10d$ iterations. It returns, for every row, the final $u$ together with its best-response scale $\sigma = \beta_t(u)/\gamma_t(u)$ of Proposition~\ref{prop:best_scale}, with $\sigma = 0$ if $\gamma_t(u) \le \varepsilon$.

\paragraph{Group update.}
The returned pair $(u, \sigma)$ replaces $(q_{i,t}, s_{i,t})$ in every row where it differs from the current pair; if no row changed its integers, the visit leaves the group unchanged. Installing the best-response scale gives $\phi_i = \mac{L}_t(u)$, so the scale of a group is re-optimized whenever the group is updated, and the row objective never increases (Proposition~\ref{prop:termination}). Because the quantities in Eq.~\eqref{eq:incr} are updated incrementally, the implementation recomputes $\phi_i$ from scratch after every group update. This guards against the accumulation of floating-point error and is not needed for the guarantees below.

\paragraph{Joint scale re-solve.}
Let $\mu_i$ be the number of groups whose pair $(q_{i,t}, s_{i,t})$ was replaced in row $i$ during a pass. After the pass, Algorithm~\ref{alg:jointq} re-solves all scales of row $i$ with its integers fixed if two conditions hold: the row is active, that is, $\phi_i$ decreased by more than $n\varepsilon$ in the previous pass (all rows are active in the first pass), and $\mu_i > \tau_{\mathrm{hi}} T$ or $\mu_i < \tau_{\mathrm{lo}} T$. It solves $B(\B{q}_i)\B{s}^\top = A(\B{q}_i)\widetilde{H} w_i^\top$, which coincides with Eq.~\eqref{eq:global_scale} when all groups of row $i$ are active. The systems of the selected rows are solved together. If any of them is singular, for example because a row has an all-zero group, all of them are re-solved with $B(\B{q}_i) + \xi_i I_T$, where $\xi_i = \max(0.01 \max_t [B(\B{q}_i)]_{tt}, 10^{-6})$. The same damped re-solve is applied to every row whose solution contains a scale with $|s_t| > 65504/L$, the bound that keeps every $s_t q_j$ representable in FP16. The new scales are installed only if they decrease $\phi_i$ by more than $n\varepsilon$; without damping, Theorem~\ref{thm:scale_elim} already ensures that the solution does not increase $\phi_i$, so this test only filters numerical noise and damped solutions. For per-channel quantization ($T = 1$), $\mu_i \in \{0, 1\}$, so every active row is selected. However, the $1 \times 1$ re-solve only recomputes the best-response scale, which the group updates and the warm-start refit of Section~\ref{sec:warm_start} have already applied, so it never passes the acceptance test. The per-channel results are therefore independent of $(\tau_{\mathrm{lo}}, \tau_{\mathrm{hi}})$. We chose $(\tau_{\mathrm{lo}}, \tau_{\mathrm{hi}}) = (0.2, 0.6)$ in a brief preliminary tuning on a few randomly selected layers, by comparing the resulting layer objective, and we have not ablated this choice.

\paragraph{Stopping rule.}
After a pass, a row is active if $\phi_i$ decreased by more than $n\varepsilon$ during the pass, that is, if the normalized objective $\phi_i/n$ decreased by more than $\varepsilon$. Algorithm~\ref{alg:jointq} stops when fewer than a fraction $\omega_{\mathrm{out}}$ of the rows are active, so the number of passes is not fixed in advance. In all experiments, $\varepsilon = 10^{-12}$ and $\omega_{\mathrm{in}} = \omega_{\mathrm{out}} = 0.1$ (Appendix~\ref{app:warm_start}).

\begin{proposition}[Descent and Termination]
    \label{prop:termination}
    Let $\lambda \ge 0$, and let $\varepsilon \ge 0$ be smaller than every positive value of $\gamma_t(u)$ over all groups $t$ and all $u \in \mac{C}_b^d$; such a threshold exists because $\mac{C}_b^d$ is finite. Then the following hold for every row $i$.
    (i) A group update either leaves $\phi_i$ unchanged or replaces it by $\mac{L}_t(u) \le \phi_i$. If Algorithm~\ref{alg:inner} applied $k \ge 1$ moves to row $i$, then $\mac{L}_t(u) < \phi_i - k\varepsilon$.
    (ii) An installed joint re-solve decreases $\phi_i$ by more than $n\varepsilon$.
    (iii) Consequently, $\phi_i$ is nonincreasing throughout Algorithm~\ref{alg:jointq}. Moreover, Algorithm~\ref{alg:inner} stops after at most $10d$ iterations, and if $\varepsilon > 0$ and $\omega_{\mathrm{out}} > 0$, Algorithm~\ref{alg:jointq} stops after at most $1 + \sum_{i} \phi_i^{(0)} / (n\, \omega_{\mathrm{out}}\, m\, \varepsilon)$ passes, where $\phi_i^{(0)}$ is the row objective of the warm start.
\end{proposition}
The proof is provided in Appendix~\ref{app:termination_proof}. The condition on $\varepsilon$ only ensures that the guard in Eq.~\eqref{eq:captured} coincides with the zero-curvature branch of Proposition~\ref{prop:best_scale}. The proposition holds for any choice of $\omega_{\mathrm{in}}$, $\tau_{\mathrm{lo}}$, and $\tau_{\mathrm{hi}}$, and the bound on the number of passes only certifies finiteness. This descent statement applies to the row objective defined by $\widetilde{H}$; when $\lambda>0$, it is a regularized objective and does not imply monotonic improvement of the unregularized output error, perplexity, or downstream accuracy. The full procedure is summarized in Algorithms~\ref{alg:jointq} and~\ref{alg:inner}.

\paragraph{Cost.}
Forming the vectors $g_{i,t}$ of all rows takes one $(m \times p)$-by-$(p \times d)$ product, that is, $\mac{O}(mpd)$ time per visit and $\mac{O}(mp^2)$ time per pass. An iteration of Algorithm~\ref{alg:inner} costs $\mac{O}(md)$ for all rows, so a visit costs $\mac{O}(md^2)$ under the cap of $10d$ iterations, and a pass costs $\mac{O}(mpd)$. Recomputing $\phi_i$ costs $\mac{O}(p^2)$ per updated row and visit, which amounts to $\mac{O}(Tmp^2)$ per pass when most rows are updated. This term dominates in group-wise settings. Since $\phi_i = \alpha_t - \kappa_t(u)$ after a group update, it could be reduced to $\mac{O}(d)$ per row and visit, but we keep the exact recomputation for numerical robustness. The joint re-solve costs $\mac{O}(p^2 + T^3)$ per selected row: forming $B(\B{q}_i)$ and the right-hand side takes $\mac{O}(p^2)$ time because each $a_t$ has at most $d$ nonzero entries, and the solve takes $\mac{O}(T^3)$ time. The number of passes is determined by the stopping rule.

\begin{algorithm}[t]
  \caption{JointQ local search for one layer}
  \label{alg:jointq}
  \small
  \begin{algorithmic}[1]
    \Require $W$, $\widetilde{H}$, groups $\{G_t\}$, codebook $\mac{C}_b$, warm start $(\B{s}_i, \B{q}_i)_{i}$; $\varepsilon$, $\omega_{\mathrm{out}}$, $\tau_{\mathrm{lo}}$, $\tau_{\mathrm{hi}}$
    \State $\phi_i \leftarrow \phi(\B{s}_i, \B{q}_i)$ for all rows $i$;\; $\mac{A} \leftarrow \{0, \dots, m{-}1\}$ \Comment{active rows}
    \Repeat
      \State $\phi_i^{\mathrm{prev}} \leftarrow \phi_i$ and $\mu_i \leftarrow 0$ for all rows $i$
      \For{$t = 0, \dots, T{-}1$} \Comment{one pass}
        \State $g_{i,t}^\top \leftarrow$ row $i$ of $(W - \widehat{W})\widetilde{H}_{:,G_t} + \widehat{W}_{:,G_t}\widetilde{H}_t$ for all rows $i$
        \State $(u_i, \sigma_i)_{i} \leftarrow$ Algorithm~\ref{alg:inner} on group $t$
        \If{$u_i \neq q_{i,t}$ for some row $i$}
          \For{each row $i$ with $(u_i, \sigma_i) \neq (q_{i,t}, s_{i,t})$}
            \State $(q_{i,t}, s_{i,t}) \leftarrow (u_i, \sigma_i)$;\; $\mu_i \leftarrow \mu_i + 1$;\; recompute $\phi_i$
          \EndFor
        \EndIf
      \EndFor
      \State $\mac{U} \leftarrow \{ i \in \mac{A} : \mu_i > \tau_{\mathrm{hi}} T \text{ or } \mu_i < \tau_{\mathrm{lo}} T \}$
      \For{each row $i \in \mac{U}$} \Comment{joint scale re-solve}
        \State $\B{s}' \leftarrow$ solution of $B(\B{q}_i)\B{s}^\top = A(\B{q}_i)\widetilde{H} w_i^\top$, with the damping fallback
        \If{$\phi(\B{s}', \B{q}_i) < \phi_i - n\varepsilon$}
          \State $\B{s}_i \leftarrow \B{s}'$;\; $\phi_i \leftarrow \phi(\B{s}', \B{q}_i)$
        \EndIf
      \EndFor
      \State $\mac{A} \leftarrow \{ i : \phi_i < \phi_i^{\mathrm{prev}} - n\varepsilon \}$
    \Until{$|\mac{A}| < \omega_{\mathrm{out}}\, m$}
    \State \Return $(\B{s}_i, \B{q}_i)_{i}$
  \end{algorithmic}
\end{algorithm}

\begin{algorithm}[t]
  \caption{Best-improvement integer search on group $t$ (all rows in parallel)}
  \label{alg:inner}
  \small
  \begin{algorithmic}[1]
    \Require $g_{i,t} \in \mab{R}^d$ and initial $u_i = q_{i,t} \in \mac{C}_b^d$ for all rows $i$; $\widetilde{H}_t$; $\varepsilon$, $\omega_{\mathrm{in}}$
    \State $c_i \leftarrow \widetilde{H}_t u_i$;\; $\beta_i \leftarrow u_i^\top g_{i,t}$;\; $\gamma_i \leftarrow u_i^\top c_i$;\; $\kappa_i \leftarrow \beta_i^2/\gamma_i$ if $\gamma_i > \varepsilon$, else $0$
    \For{$\ell = 1, \dots, 10d$}
      \For{each row $i$, coordinate $j$, and $\delta \in \{+1, -1\}$}
        \State $\beta_{ij}^{\delta} \leftarrow \beta_i + \delta [g_{i,t}]_j$;\; $\gamma_{ij}^{\delta} \leftarrow \gamma_i + 2\delta [c_i]_j + [\widetilde{H}_t]_{jj}$
        \State $\kappa_{ij}^{\delta} \leftarrow (\beta_{ij}^{\delta})^2/\gamma_{ij}^{\delta}$ if $u_{ij} + \delta \in \mac{C}_b$ and $\gamma_{ij}^{\delta} > \varepsilon$, else $0$
      \EndFor
      \State $\kappa_i^{\delta} \leftarrow \max_j \kappa_{ij}^{\delta}$ and $j_i^{\delta} \in \argmax_j \kappa_{ij}^{\delta}$ for $\delta = \pm 1$, for all rows $i$
      \State $\delta_i \leftarrow +1$ if $\kappa_i^{+} \ge \kappa_i^{-}$, else $-1$;\; $\Delta_i \leftarrow \kappa_i^{\delta_i} - \kappa_i$ for all rows $i$
      \If{$|\{ i : \Delta_i < \varepsilon \}| > \omega_{\mathrm{in}}\, m$}
        \State \textbf{break}
      \EndIf
      \For{each row $i$ with $\Delta_i > \varepsilon$} \Comment{one move per row}
        \State $j \leftarrow j_i^{\delta_i}$;\; $u_{ij} \leftarrow u_{ij} + \delta_i$;\; $(\beta_i, \gamma_i, \kappa_i) \leftarrow (\beta_{ij}^{\delta_i}, \gamma_{ij}^{\delta_i}, \kappa_i^{\delta_i})$;\; $c_i \leftarrow c_i + \delta_i [\widetilde{H}_t]_{:,j}$
      \EndFor
    \EndFor
    \State Recompute $\beta_i$, $\gamma_i$ from $u_i$;\; $\sigma_i \leftarrow \beta_i/\gamma_i$ if $\gamma_i > \varepsilon$, else $0$
    \State \Return $(u_i, \sigma_i)_{i}$
  \end{algorithmic}
\end{algorithm}

\subsection{Tikhonov Regularization}
\label{sec:regularization}
The Hessian matrix $H = XX^\top$ can be ill-conditioned when the calibration set is small relative to $p$.
To enhance robustness while prioritizing input features that are important on the calibration distribution, JointQ uses a diagonal Tikhonov regularizer. Define the activation energy of input feature $j$ and its average across features as
\begin{equation}
  d_j \coloneqq H_{jj} = \sum_{i=1}^n X_{ji}^2,
  \qquad
  \bar d \coloneqq \frac{1}{p}\sum_{j=1}^p d_j
  = \frac{\tr(H)}{p}.
  \label{eq:activation_energy}
\end{equation}
Because the activations need not be centered, $d_j/n$ is a noncentral second moment rather than a variance. For $\bar d>0$, we set
\begin{equation}
  \rho_j \coloneqq \left(\frac{d_j}{\bar d}\right)^\eta,
  \qquad
  R \coloneqq \operatorname{diag}(\rho_1,\dots,\rho_p),
  \qquad \eta = 0.5,
  \label{eq:diag_regularizer}
\end{equation}
and replace $H$ with
\begin{equation}
  \widetilde{H} \coloneqq H + n\lambda R.
  \label{eq:reg_hessian}
\end{equation}
The exponent $\eta$ controls how strongly the penalty on feature $j$ scales with its activation energy: $\eta = 0$ gives a uniform Tikhonov term $R = I_p$, and $\eta = 1$ makes $\rho_j$ proportional to $d_j$. We fix $\eta = 0.5$ in all experiments and have not tuned it (Appendix~\ref{app:reg_settings}). Importantly, ``diagonal'' refers only to the added matrix $R$; the off-diagonal entries of $H$, which encode correlations between input features, remain unchanged.

This substitution is equivalent to solving
\begin{equation}
  \min_{\B{s}, \B{q}}
  \lVert WX - \widehat{W} X \rVert_F^2
  + n\lambda \lVert (W - \widehat{W}) R^{1/2} \rVert_F^2,
  \label{eq:reg_obj}
\end{equation}
whose regularization term can be written as
\begin{equation}
  n\lambda \sum_{j=1}^p \rho_j
  \lVert W_{:,j} - \widehat{W}_{:,j} \rVert_2^2.
  \label{eq:reg_columnwise}
\end{equation}
Thus, the regularizer does not shrink $\widehat{W}$ toward zero; it discourages quantization errors relative to the original weights, with stronger absolute penalties on features having larger calibration activation energy. The factor $n$ makes $\lambda$ interpretable relative to the normalized Hessian $\tfrac{1}{n}XX^\top$. The equivalence is proved in Appendix~\ref{app:reg_proof}, and Appendix~\ref{app:reg_corollary} gives the resulting weighted curvature and scale-sensitivity bounds.

Equation~\eqref{eq:reg_hessian} does not necessarily make the Hessian positive definite: if $d_j=0$, then $\rho_j=0$ and no regularization is added in that direction. We therefore state $\widetilde{H}\succ0$ explicitly in results that require uniqueness or a matrix inverse. The implementation also supports positive-semidefinite cases through the canonical zero-curvature branch in Proposition~\ref{prop:best_scale}.

\subsection{Multi-Candidate Warm Start}
\label{sec:warm_start}
Because the search in Section~\ref{sec:local_search} is local, its result depends on the initial state. JointQ therefore builds its warm start from a pool of candidate solutions and selects among them independently for each output row. Every step below uses the same regularized Hessian $\widetilde{H}$ as the subsequent local search. All experiments in this paper use the two candidates described next; additional candidates can be added to the pool without changing the procedure.

\paragraph{Clip-Optimize candidate.}
For each row and group, we set the initial scale to the symmetric min--max value $s_t^{(0)} = \max_{j \in G_t} |w_j| / (L-1)$ and the integers to $q_j^{(0)} = \mathrm{clamp}(\lfloor w_j / s_t^{(0)} \rceil, -L, L-1)$ with round-half-to-even, where $s_t^{(0)}$ is clamped below at $10^{-8}$, so that an all-zero group starts from $q_t^{(0)} = 0$. Unlike Eq.~\eqref{eq:scale_minmax}, this rule maps the entry of largest magnitude exactly to $\pm(L-1)$. We then refine each group by a single call of Algorithm~\ref{alg:inner} on the single-group problem
\begin{equation}
  \min_{\sigma \in \mab{R},\, u \in \mac{C}_b^d}
  \bigl(w_{G_t}^\top - \sigma u\bigr)^\top \widetilde{H}_t \bigl(w_{G_t}^\top - \sigma u\bigr),
  \label{eq:group_local}
\end{equation}
where $w_{G_t} \in \mab{R}^{1 \times d}$ is the restriction of $w$ to $G_t$ and $\widetilde{H}_t$ is the principal submatrix of Section~\ref{sec:error_decomp}. That is, Algorithm~\ref{alg:inner} is run with $g_{i,t} = \widetilde{H}_t w_{i,G_t}^\top$, the clipped integers as initial values, and the same $\varepsilon$, $\omega_{\mathrm{in}}$, and iteration cap as in Algorithm~\ref{alg:jointq}, and the returned pair $(u, \sigma)$ becomes the integers and scale of the group. This group-local objective ignores the off-diagonal blocks of $\widetilde{H}$, so the groups can be refined independently of one another. The candidate is thus a direct group-wise extension of scale-eliminated per-channel search that treats the groups independently; the local search of Section~\ref{sec:local_search}, which starts from the selected candidate, accounts for the cross-group coupling that this candidate ignores. For $T = 1$, $\widetilde{H}_t = \widetilde{H}$ and Eq.~\eqref{eq:group_local} is the full row objective, so the candidate is a Beacon-type search~\citep{zhang2025beacon} started from round-to-nearest, with $\pm 1$ best-improvement moves in place of Beacon's greedy initialization and cyclic coordinate updates.

\paragraph{GPTQ candidate.}
We run GPTQ with the same bit-width, grouping, and codebook, using $\widetilde{H}$ in place of $H$ and the same damping $\epsilon = 0.01$ as the GPTQ baseline (Section~\ref{sec:gptq_background}). Unless otherwise noted, this run uses min--max scales and no activation ordering, so for $\lambda = 0$ its integers coincide with those of the GPTQ baseline. The variant in which this candidate is produced by \gptqstar is described in Appendix~\ref{app:gptqstar}.

\paragraph{Global scale refit.}
For each candidate $c$, we keep the integers $\B{q}_i^{(c)}$ of every row $i$ fixed and re-solve all group scales by the joint scale re-solve of Section~\ref{sec:local_search}, applied to all rows and including its damping fallback and acceptance test. We write $\phi_i^{(c)}$ for the objective of row $i$ of candidate $c$ after this refit. Because the refit installs new scales only if they decrease the objective, $\phi_i^{(c)}$ is at most the objective before the refit; because the refitted scales are feasible for $\B{q}_i^{(c)}$, Theorem~\ref{thm:scale_elim} gives $\phi_i^{(c)} \ge F_i(\B{q}_i^{(c)})$, where $F_i$ denotes Eq.~\eqref{eq:reduced_obj} for row $w_i$. When the undamped solution is installed, the refitted scales are the jointly optimal scales $\B{s}_I^\star(\B{q}_i^{(c)})$ of Eq.~\eqref{eq:global_scale}, and equality holds. The refitted scales may be negative (Section~\ref{sec:formulation}); in particular, the scales of the GPTQ candidate generally differ from those produced by GPTQ. As in Section~\ref{sec:local_search}, the refit costs $\mac{O}(p^2 + T^3)$ per row.

\paragraph{Row-wise selection.}
For each output row $i$, we select $c_i^\star \in \argmin_c \phi_i^{(c)}$, preferring the Clip-Optimize candidate on ties, and initialize Algorithm~\ref{alg:jointq} with the refitted scales and integers of candidate $c_i^\star$. Different rows of a layer may therefore start from different candidates, whereas all groups of a row come from a single candidate. Selection uses only the layer-local objective on the calibration data and never accesses evaluation data or downstream metrics. Algorithm~\ref{alg:warm_start} in Appendix~\ref{app:warm_start} summarizes the procedure.

Because the pool contains the GPTQ candidate, the warm start yields the following guarantee.

\begin{proposition}[Dominance over the GPTQ candidate]
\label{prop:warm_dominance}
Under the assumptions of Proposition~\ref{prop:termination}, for each row $i$, let $(\B{s}_i^{\mathrm{G}}, \B{q}_i^{\mathrm{G}})$ be the GPTQ candidate before the scale refit, let $\phi_i^{(\mathrm{G})}$ be its objective after the refit, and let $(\B{s}_i^{\mathrm{J}}, \B{q}_i^{\mathrm{J}})$ be the output of Algorithm~\ref{alg:jointq} started from the warm start above. Then
\begin{equation}
  \phi(\B{s}_i^{\mathrm{J}}, \B{q}_i^{\mathrm{J}})
  \le \phi_i^{(c_i^\star)}
  \le \phi_i^{(\mathrm{G})}
  \le \phi(\B{s}_i^{\mathrm{G}}, \B{q}_i^{\mathrm{G}}),
  \label{eq:warm_dominance}
\end{equation}
irrespective of the stopping rule and the thresholds of Algorithm~\ref{alg:jointq}. If, in addition, $\widetilde{H} \succ 0$, then $F_i(\B{q}_i^{\mathrm{G}}) \le \phi_i^{(\mathrm{G})}$, with equality when the undamped refit is installed.
\end{proposition}
The proof is given in Appendix~\ref{app:warm_dominance_proof}. For $\lambda = 0$, the GPTQ candidate has the same integers as the GPTQ baseline, so Eq.~\eqref{eq:warm_dominance} states that unregularized JointQ never has a larger row-wise calibration output error than GPTQ. The guarantee rests on including the GPTQ solution in the pool together with the monotonicity of the local search (Proposition~\ref{prop:termination}); methods that construct the integers without reference to GPTQ have no such guarantee. For $\lambda > 0$, the inequality holds for the regularized objective and relative to GPTQ run on $\widetilde{H}$, not relative to the GPTQ baseline; it says nothing about perplexity or downstream accuracy.

\subsection{Implementation}
\label{sec:implementation}
We use the diagonal regularizer in Eq.~\eqref{eq:diag_regularizer} with the fixed exponent $\eta=0.5$. The regularization strength is set per quantization setting and layer type and shared across all model families; Section~\ref{sec:experiments} summarizes the values and how they were selected, and Appendix~\ref{app:reg_settings} gives the full table and selection procedure. When $\lambda>0$, the same regularized Hessian is used in every step of the warm start (both candidates, the scale refit, and the row-wise selection) and in the local search. Unless otherwise noted, a layer assigned $\lambda=0$ is first quantized without regularization. If its relative weight error $\lVert W-\widehat W\rVert_F^2/\lVert W\rVert_F^2$ is at least $0.2$, the entire layer quantization is retried with $\lambda=0.01$: both candidates are regenerated, the scales are refitted, the selection is repeated, and the local search is rerun, all with the new regularized Hessian. The retry is therefore not a post-hoc correction of the scales. The retry strength $0.01$ is the smallest nonzero value in our candidate grid, and the threshold $0.2$ was set from preliminary W4G128 runs (Appendix~\ref{app:reg_settings}). Because the retry applies only to layers assigned $\lambda = 0$, it is never triggered in a setting in which every layer type has $\lambda > 0$.

\paragraph{Zero point and storage.}
In the fixed-zero-point format, every group uses $z_t = L = 2^{b-1}$, exactly as in GPTQ. JointQ modifies only the centered integers $q_t \in \mac{C}_b^d$ and the scales $s_t \in \mab{R}$, so its output is packed in the same way as a GPTQ checkpoint: raw integers $q_j + L \in \{0, \ldots, 2^b - 1\}$ and one scale per group, stored in FP16.

\paragraph{Numerical settings.}
All computations of JointQ outside the GPTQ run that produces the GPTQ candidate are performed in float64. The tolerances and thresholds of Algorithms~\ref{alg:jointq}--\ref{alg:warm_start} are fixed in all experiments and listed in Table~\ref{tab:search_settings} of Appendix~\ref{app:warm_start}.
\section{Experiments}
\label{sec:experiments}

We empirically evaluate JointQ, investigating its overall effectiveness and the contribution of regularization.

\paragraph{Setup.}
We evaluate JointQ on four models from two families: Llama-2-7B~\citep{touvron2023llama2}, Llama-3-8B~\citep{dubey2024llama3}, Qwen3-8B, and Qwen3-14B~\citep{yang2025qwen3}.\footnote{All four models are pretrained base checkpoints from Hugging Face, so Qwen3-8B and Qwen3-14B refer to the base models rather than the post-trained models: \nolinkurl{meta-llama/Llama-2-7b-hf}, \nolinkurl{meta-llama/Meta-Llama-3-8B}, \nolinkurl{Qwen/Qwen3-8B-Base}, and \nolinkurl{Qwen/Qwen3-14B-Base}.}
The main evaluation uses weight-only symmetric quantization in W4G128, W4 per-channel, W3G128, and W3 per-channel settings. For calibration, we sample 1024 sequences of length 2048 tokens from the C4 training split~\citep{raffel2020exploring}, using seed 0. The unquantized models use bfloat16 weights. We report perplexity on WikiText-2~\citep{merity2016pointer}. Zero-shot performance is measured using lm-eval-harness~\citep{eval-harness} on ARC-c, ARC-e~\citep{clark2018arc}, PIQA~\citep{bisk2020piqa}, WinoGrande~\citep{sakaguchi2021winogrande}, HellaSwag~\citep{zellers2019hellaswag}, and BoolQ~\citep{clark2019boolq}. We use normalized accuracy where available and raw accuracy for WinoGrande and BoolQ; ``Avg.'' is the arithmetic mean over all six tasks. JointQ uses the diagonal regularizer in Eq.~\eqref{eq:diag_regularizer} with $\eta = 0.5$ and, in the four main settings, $\lambda \in \{0, 0.01, 0.15\}$, set per setting and layer type and shared across all model families; the larger value $0.15$ appears only in the per-channel settings, and W3 per-channel assigns $\lambda > 0$ to every layer type (Table~\ref{tab:layerwise_lambda} in Appendix~\ref{app:reg_settings}). These values were chosen by a manual preliminary search on Llama-3-8B, Qwen3-8B, and Qwen3-14B that used WikiText-2 perplexity and zero-shot accuracy on ARC-c, ARC-e, PIQA, and WinoGrande, that is, evaluation data also used for the results below; Llama-2-7B, HellaSwag, and BoolQ were not used in this search (Appendix~\ref{app:reg_settings}). For the four main settings, entries with $\lambda=0$ use the retry rule described in Section~\ref{sec:implementation}. Table~\ref{tab:ablation_reg} separates the contribution of the unregularized JointQ pipeline from that of the full stabilization policy. Additional comparisons with GPTQ scale-search and activation-ordering heuristics are provided in Appendix~\ref{app:gptqstar}. Throughout, \textbf{JointQ} denotes the complete configuration: the multi-candidate warm start of Section~\ref{sec:warm_start} with Clip-Optimize and vanilla GPTQ candidates, the layer-specific regularization in Table~\ref{tab:layerwise_lambda}, and adaptive retry. \textbf{JointQ w/o stab.} disables both regularization and retry, so the entire pipeline, including warm-start generation and selection, runs with $\lambda=0$. \textbf{JointQ (\gptqstar cand.)} replaces the vanilla GPTQ candidate with a \gptqstar candidate while retaining the Clip-Optimize candidate; it is therefore not a pure \gptqstar warm start. Every JointQ result in this paper, including the W2G32 and calibration-sensitivity experiments, uses this two-candidate warm start. The tolerances and thresholds of the local search are fixed in all experiments (Table~\ref{tab:search_settings} in Appendix~\ref{app:warm_start}). Baselines are \textbf{GPTQ} (min--max scales, no activation ordering), \textbf{GPTQ-MSE} (MSE-based scale search only), and \textbf{\gptqstar} (MSE-based scale search and activation ordering; Appendix~\ref{app:gptqstar}).
All methods use the same fixed-zero-point codebook $\mac{C}_b$ and zero point $2^{b-1}$ (Section~\ref{sec:group_format}), matching the GPTQ reference implementation. JointQ and all baselines are implemented in the open-source OneComp library~\citep{ichikawa2026onecomp}, and every result in this paper was obtained with OneComp v1.4.2 (Appendix~\ref{app:warm_start}).

\paragraph{Hardware and timing.}
Each experiment is allocated a single NVIDIA B200 GPU, 16 CPU cores, and 240\,GB of host memory; all nodes used have the same configuration. Reported quantization times are the total wall-clock seconds spent quantizing the linear layers and exclude the calibration forward passes used to collect $H$; for JointQ, they include warm-start generation and any adaptive retries. Each node hosts eight GPUs, and other jobs may have run on the same node concurrently, so the times are indicative rather than precise benchmarks. Since JointQ performs all computations outside its GPTQ candidate run in float64 (Section~\ref{sec:implementation}), its time relative to GPTQ depends on the FP64 throughput of the GPU and can be considerably larger on GPUs with limited FP64 support.

\subsection{Main Results}
\label{subsec:main_results}

Table~\ref{tab:main_results} compares JointQ with GPTQ across four quantization regimes that vary in bit-width and grouping granularity. The GPTQ baseline and the GPTQ candidate inside JointQ both use \texttt{actorder=false} and \texttt{mse=false}. These settings cover the standard W4G128 configuration and more challenging per-channel and 3-bit configurations. Proposition~\ref{prop:separation} provides an idealized mechanism by which frozen-scale assignment can become more restrictive with smaller bit-widths or larger groups; it does not imply monotonic improvement of downstream metrics. We evaluate whether the empirical trend is consistent with this mechanism.

\begin{table*}[t]
  \centering\small\setlength{\tabcolsep}{3.5pt}
  \caption{WikiText-2 perplexity ($\downarrow$) and average zero-shot accuracy ($\uparrow$). JointQ uses the layer-specific diagonal regularization in Table~\ref{tab:layerwise_lambda} (Appendix~\ref{app:reg_settings}). Bold denotes the better quantized result for each model and setting.}
  \label{tab:main_results}
  \begin{tabular}{ll rr rr rr rr}
    \toprule
    & & \multicolumn{2}{c}{W4G128} & \multicolumn{2}{c}{W4 per-ch} & \multicolumn{2}{c}{W3G128} & \multicolumn{2}{c}{W3 per-ch} \\
    \cmidrule(lr){3-4} \cmidrule(lr){5-6} \cmidrule(lr){7-8} \cmidrule(lr){9-10}
    Model & Method & PPL & Avg. & PPL & Avg. & PPL & Avg. & PPL & Avg. \\
    \midrule
    \multirow{3}{*}{Llama-2-7B}
    & BF16 & \multicolumn{8}{c}{PPL: 5.47 \quad Avg.: 0.7038} \\
    & GPTQ & 5.89 & \textbf{0.6918} & 6.17 & 0.6816 & 6.80 & 0.6359 & 27.74 & 0.5020 \\
    \rowcolor{gray!12} & JointQ & \textbf{5.67} & 0.6914 & \textbf{5.96} & \textbf{0.6836} & \textbf{6.78} & \textbf{0.6610} & \textbf{7.98} & \textbf{0.6032} \\
    \midrule
    \multirow{3}{*}{Llama-3-8B}
    & BF16 & \multicolumn{8}{c}{PPL: 6.14 \quad Avg.: 0.7442} \\
    & GPTQ & 14.09 & 0.7248 & 399.02 & 0.5960 & 57.69 & 0.5036 & 3381.46 & 0.3732 \\
    \rowcolor{gray!12} & JointQ & \textbf{6.65} & \textbf{0.7426} & \textbf{7.41} & \textbf{0.7247} & \textbf{8.75} & \textbf{0.7060} & \textbf{13.09} & \textbf{0.6606} \\
    \midrule
    \multirow{3}{*}{Qwen3-8B}
    & BF16 & \multicolumn{8}{c}{PPL: 9.73 \quad Avg.: 0.7403} \\
    & GPTQ & 10.33 & 0.7323 & 11.04 & 0.7114 & 11.75 & 0.6919 & \textbf{19.55} & 0.5449 \\
    \rowcolor{gray!12} & JointQ & \textbf{10.17} & \textbf{0.7336} & \textbf{10.71} & \textbf{0.7240} & \textbf{11.50} & \textbf{0.7250} & 22.11 & \textbf{0.6686} \\
    \midrule
    \multirow{3}{*}{Qwen3-14B}
    & BF16 & \multicolumn{8}{c}{PPL: 8.65 \quad Avg.: 0.7736} \\
    & GPTQ & \textbf{8.85} & \textbf{0.7709} & \textbf{9.11} & 0.7525 & 10.07 & 0.7356 & \textbf{13.84} & 0.6089 \\
    \rowcolor{gray!12} & JointQ & 8.89 & 0.7699 & 9.40 & \textbf{0.7710} & \textbf{9.91} & \textbf{0.7472} & 18.92 & \textbf{0.7203} \\
    \bottomrule
  \end{tabular}
\end{table*}

JointQ improves average zero-shot accuracy over GPTQ in 14 of the 16 model--configuration pairs and individual task accuracy in 80 of 96 comparisons, with one tie. Averaged over the four models, the accuracy gains are $0.0044$ for W4G128, $0.0405$ for W4 per-channel, $0.0680$ for W3G128, and $0.1559$ for W3 per-channel. This progression is consistent with the mechanism in Proposition~\ref{prop:separation}: reducing the bit-width limits the available integer directions, while per-channel quantization couples more weights to a single scale, making frozen-scale assignment increasingly restrictive. The two average-accuracy decreases, $0.0004$ on Llama-2-7B and $0.0010$ on Qwen3-14B in W4G128, are small.

The advantage is particularly pronounced on Llama-3-8B, where JointQ improves both metrics in every setting and mitigates the severe degradation of GPTQ in the per-channel regimes. For example, in W3 per-channel, JointQ reduces perplexity from $3381.46$ to $13.09$ and raises average accuracy from $0.3732$ to $0.6606$. The trend is not confined to one model: in W3 per-channel, the accuracy gains are $0.1012$ on Llama-2-7B, which was not used to select $\lambda$, $0.1237$ on Qwen3-8B, and $0.1114$ on Qwen3-14B.

Perplexity follows a less uniform pattern: JointQ improves it in 12 of the 16 comparisons but worsens it in four Qwen3 cases. The clearest discrepancy occurs for Qwen3-14B W3 per-channel, where perplexity increases from $13.84$ to $18.92$ despite the accuracy improvement from $0.6089$ to $0.7203$. Thus, the results do not support uniform dominance on every metric. Instead, JointQ better preserves downstream accuracy as quantization becomes more aggressive, while its effect on language-modeling perplexity can be model dependent.

\paragraph{Comparison with MSE-based scale search.}
To test whether JointQ's gains can be reproduced by improving only the initial scales while keeping them fixed during integer assignment, we compare JointQ with GPTQ-MSE. GPTQ-MSE enables MSE-based scale search but not activation ordering, so both configurations are evaluated under an actorder-free condition. Table~\ref{tab:mse_summary} summarizes the model-level comparisons; full results are reported in Appendix~\ref{app:gptqstar}. JointQ achieves higher average accuracy in 15 of 16 comparisons and lower perplexity in 12, with one tie. In G128, it wins 7 of 8 accuracy comparisons and 6 of 8 perplexity comparisons; its W3G128 accuracy gain is positive for all four models. These results indicate that the advantage of JointQ, particularly in W3G128, cannot be explained solely by the min--max scale initialization used by vanilla GPTQ. Note, however, that JointQ itself refits all group scales with the integers fixed, both in its warm start (Section~\ref{sec:warm_start}) and, for selected rows, after each pass of the local search (Section~\ref{sec:local_search}); these refits are Hessian-aware scale improvements. This comparison therefore does not separate the contribution of the fixed-integer refits from that of the integer moves.

\begin{table}[t]
  \centering\small\setlength{\tabcolsep}{5pt}
  \caption{JointQ versus GPTQ-MSE. Entries count model-level wins/ties/losses for JointQ; $\Delta$ Avg. is the mean JointQ accuracy gain.}
  \label{tab:mse_summary}
  \begin{tabular}{l rrr}
    \toprule
    Setting & PPL W/T/L & Avg. W/L & $\Delta$ Avg. \\
    \midrule
    W4G128 & 3/0/1 & 3/1 & $+0.0029$ \\
    W4 per-ch & 3/1/0 & 4/0 & $+0.0316$ \\
    W3G128 & 3/0/1 & 4/0 & $+0.0257$ \\
    W3 per-ch & 3/0/1 & 4/0 & $+0.1435$ \\
    \midrule
    Overall & 12/1/3 & 15/1 & -- \\
    \bottomrule
  \end{tabular}
\end{table}

\subsection{Ablation: Unregularized Pipeline and Stabilization}
\label{subsec:local_stabilization}
Table~\ref{tab:ablation_reg} provides a staged comparison across all four models. Comparing GPTQ with JointQ w/o stab. measures the effect of the unregularized JointQ pipeline, that is, the multi-candidate warm start together with the local search; this comparison does not separate the two components. Comparing JointQ w/o stab. with JointQ measures the full stabilization policy, including its effect on the warm start, rather than regularization or retry separately.

\begin{table*}[t]
  \centering\scriptsize\setlength{\tabcolsep}{3.5pt}
  \caption{Staged ablation of the unregularized JointQ pipeline and stabilization. Bold denotes the best result in each row and metric.}
  \label{tab:ablation_reg}
  \begin{tabular}{ll rr rr rr}
    \toprule
    & & \multicolumn{2}{c}{GPTQ} & \multicolumn{2}{c}{JointQ w/o stab.} & \multicolumn{2}{c}{JointQ} \\
    \cmidrule(lr){3-4} \cmidrule(lr){5-6} \cmidrule(lr){7-8}
    Model & Setting & PPL$\downarrow$ & Avg.$\uparrow$ & PPL$\downarrow$ & Avg.$\uparrow$ & PPL$\downarrow$ & Avg.$\uparrow$ \\
    \midrule
    \multirow{4}{*}{Llama-2-7B}
    & W4G128 & 5.89 & \textbf{0.6918} & \textbf{5.67} & 0.6896 & \textbf{5.67} & 0.6914 \\
    & W4 per-ch & 6.17 & 0.6816 & 6.27 & 0.6694 & \textbf{5.96} & \textbf{0.6836} \\
    & W3G128 & 6.80 & 0.6359 & 6.89 & 0.6485 & \textbf{6.78} & \textbf{0.6610} \\
    & W3 per-ch & 27.74 & 0.5020 & 23.02 & 0.4511 & \textbf{7.98} & \textbf{0.6032} \\
    \midrule
    \multirow{4}{*}{Llama-3-8B}
    & W4G128 & 14.09 & 0.7248 & \textbf{6.65} & 0.7348 & \textbf{6.65} & \textbf{0.7426} \\
    & W4 per-ch & 399.02 & 0.5960 & 7.59 & \textbf{0.7256} & \textbf{7.41} & 0.7247 \\
    & W3G128 & 57.69 & 0.5036 & 8.90 & 0.6905 & \textbf{8.75} & \textbf{0.7060} \\
    & W3 per-ch & 3381.46 & 0.3732 & 315.46 & 0.6324 & \textbf{13.09} & \textbf{0.6606} \\
    \midrule
    \multirow{4}{*}{Qwen3-8B}
    & W4G128 & 10.33 & 0.7323 & \textbf{10.11} & 0.7305 & 10.17 & \textbf{0.7336} \\
    & W4 per-ch & 11.04 & 0.7114 & \textbf{10.66} & 0.7207 & 10.71 & \textbf{0.7240} \\
    & W3G128 & 11.75 & 0.6919 & 11.54 & 0.7186 & \textbf{11.50} & \textbf{0.7250} \\
    & W3 per-ch & \textbf{19.55} & 0.5449 & 30.19 & \textbf{0.6697} & 22.11 & 0.6686 \\
    \midrule
    \multirow{4}{*}{Qwen3-14B}
    & W4G128 & \textbf{8.85} & \textbf{0.7709} & 8.88 & \textbf{0.7709} & 8.89 & 0.7699 \\
    & W4 per-ch & \textbf{9.11} & 0.7525 & 9.41 & 0.7684 & 9.40 & \textbf{0.7710} \\
    & W3G128 & 10.07 & 0.7356 & \textbf{9.91} & \textbf{0.7479} & \textbf{9.91} & 0.7472 \\
    & W3 per-ch & \textbf{13.84} & 0.6089 & 20.20 & \textbf{0.7267} & 18.92 & 0.7203 \\
    \bottomrule
  \end{tabular}
\end{table*}

\begin{figure*}[t]
  \centering
  \includegraphics[width=\textwidth]{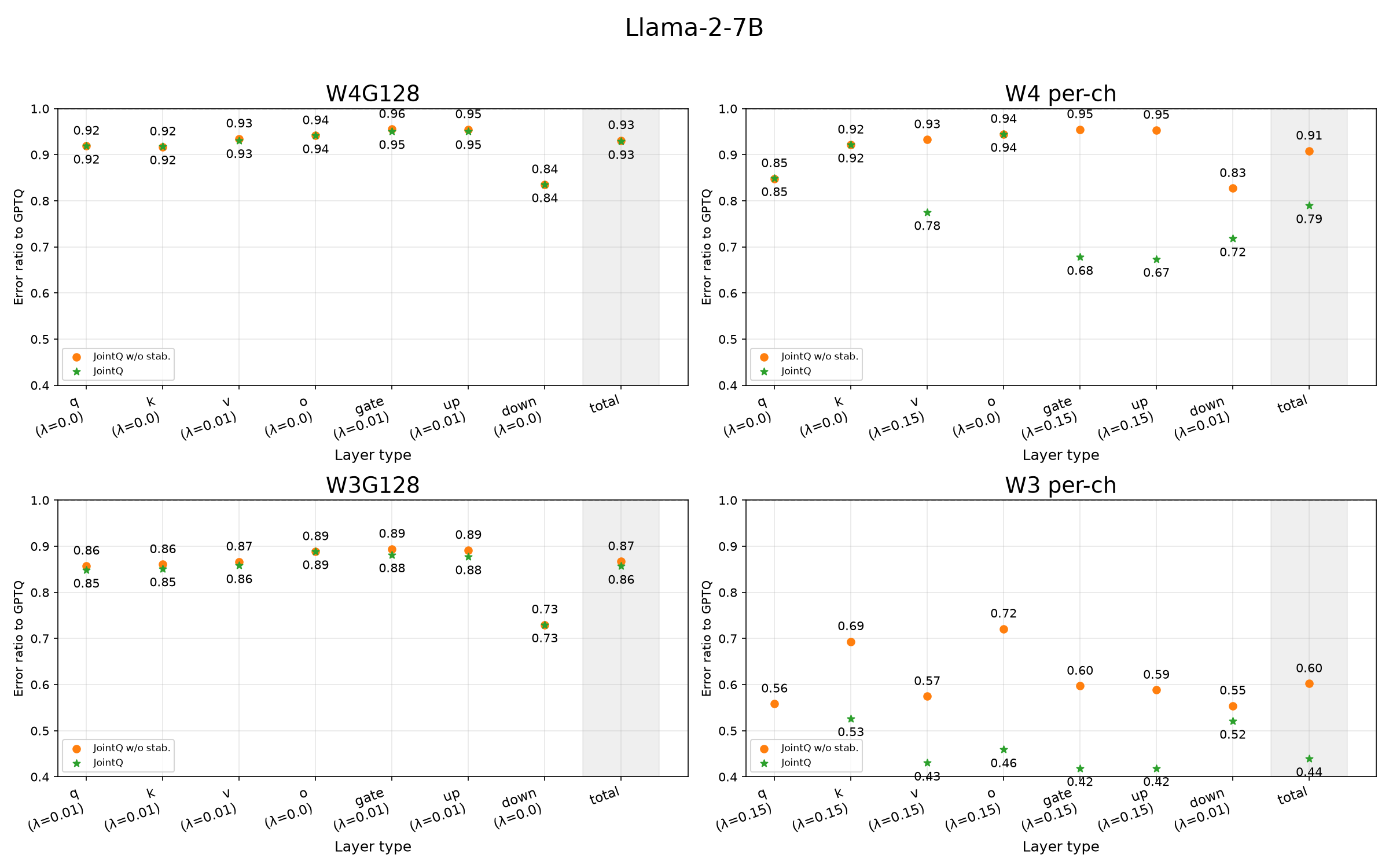}
  \caption{Layer-type-wise unregularized output-error ratios on the common calibration inputs for Llama-2-7B. At each layer, $X$ is the BF16-model calibration activation used by all methods and to form $H=XX^\top$ for quantization. For each layer type, $\lVert(W-\widehat W)X\rVert_F^2$ is summed over all transformer blocks and normalized by the corresponding GPTQ sum; ``total'' sums all seven layer types before normalization. A ratio below one indicates a lower error than GPTQ.}
  \label{fig:layer_error_llama2}
\end{figure*}

Figure~\ref{fig:layer_error_llama2} reports unregularized calibration output error; Appendix~\ref{app:layer_error} provides the other three models. At each layer, $X$ is the BF16-model calibration activation used by all methods and to form $H=XX^\top$ for quantization. Across every model, quantization setting, and layer type shown, JointQ w/o stab. has an error ratio below one. This sign is guaranteed by construction rather than an empirical finding: with $\lambda=0$, the GPTQ candidate has the same integers as the GPTQ baseline, so Proposition~\ref{prop:warm_dominance} implies that no row of JointQ w/o stab. has a larger unregularized calibration error than GPTQ. The figures therefore quantify the size of the reduction, not its sign.

The layer-wise error reduction does not, however, predict downstream improvement. Comparing GPTQ with JointQ w/o stab. in Table~\ref{tab:ablation_reg}, the unregularized pipeline improves perplexity in 10 of 16 model--configuration pairs and worsens it in 6; average accuracy improves in 11, worsens in 4, and ties in 1. On Llama-3-8B W4G128, a total error ratio of $0.92$ (Figure~\ref{fig:layer_error_llama3}) accompanies a large perplexity reduction from $14.09$ to $6.65$. In contrast, Llama-2-7B W4 per-channel has a similar ratio of $0.91$ (Figure~\ref{fig:layer_error_llama2}), yet perplexity worsens from $6.17$ to $6.27$ and average accuracy decreases from $0.6816$ to $0.6694$. On Qwen3-14B W3 per-channel, the error falls to $0.66$ of GPTQ (Figure~\ref{fig:layer_error_qwen14}) while perplexity worsens from $13.84$ to $20.20$. Thus, even a substantial reduction of the layer-local calibration objective need not translate into better end-to-end metrics.

The figures also show that, for layer types assigned $\lambda>0$, JointQ often attains lower unregularized calibration error than JointQ w/o stab.; layer types with $\lambda=0$ coincide unless adaptive retry is triggered. The gap for $\lambda>0$ is larger on the Llama models and smaller on the Qwen3 models; in particular, the Qwen3 query, key, and value projections remain close even with $\lambda=0.15$ in W3 per-channel. This is notable because the regularized objective adds a nonnegative penalty and does not guarantee improvement in the unregularized error. It suggests that regularization may lead the quantization procedure to a different and better solution of the unregularized objective. However, the same regularized Hessian is used for both warm-start candidates, the scale refit, the row-wise selection, and the subsequent local search, so these results do not isolate whether the difference arises from the warm start or from the local-search trajectory.

The downstream effect of stabilization is most pronounced in the constrained W3 per-channel regime. Relative to JointQ w/o stab., JointQ reduces perplexity from $23.02$ to $7.98$ on Llama-2-7B and from $315.46$ to $13.09$ on Llama-3-8B. It also improves W3 per-channel perplexity from $30.19$ to $22.11$ on Qwen3-8B and from $20.20$ to $18.92$ on Qwen3-14B. Because every layer type is assigned $\lambda > 0$ in W3 per-channel, adaptive retry is never triggered in this setting, so these differences are due to regularization alone. They support the role of regularization in mitigating severe perplexity degradation, including on Llama-2-7B, which was not used to select $\lambda$.

JointQ does not uniformly outperform JointQ w/o stab. across models and metrics. Over all 16 model--configuration pairs, JointQ improves perplexity in 10 cases, ties in 3, and worsens it in 3; average accuracy improves in 11 cases and worsens in 5. In particular, W3 per-channel accuracy decreases slightly on both Qwen models despite the perplexity improvements. Because a retry requantizes an entire layer, adaptive retry can also add substantial offline quantization cost in the settings where it can be triggered, namely W4G128, W4 per-channel, and W3G128.

\subsection{Calibration-set Sensitivity}
\label{subsec:calibration_sensitivity}
We evaluate calibration-set sensitivity on Llama-3-8B in W3G128 and W4G128 using 8--128 length-2048 calibration sequences and 20 paired seeds per sequence count. Full results are reported in Appendix~\ref{app:calibration_stability}.

In W3G128, JointQ has lower PPL variance than \gptqstar at every sequence count; with eight sequences, the results are $11.10 \pm 0.30$ and $11.37 \pm 1.67$, respectively. The full stabilization policy also lowers PPL relative to JointQ w/o stab. in all 100 W3G128 runs and 99 of 100 W4G128 runs. JointQ (\gptqstar cand.) lowers W4G128 PPL in 94 of 100 runs without a consistent accuracy difference, whereas W3G128 exhibits an accuracy--perplexity trade-off. These results demonstrate robust low-sample behavior for JointQ and a strong contribution from its stabilization policy.

\subsection{Additional Results in the 2-bit Regime}
\label{subsec:two_bit}
Table~\ref{tab:w2g32} reports W2G32 quantization results on all four models. JointQ uses the W2G32 layer-specific regularization values in Table~\ref{tab:layerwise_lambda} with adaptive retry disabled, because with the threshold and retry strength of Section~\ref{sec:implementation} it degraded both perplexity and accuracy in preliminary runs (Appendix~\ref{app:reg_settings}). We report this aggressive setting separately from the main results because its absolute model quality remains far below BF16 despite substantial improvements over GPTQ.

\begin{table*}[t]
  \centering\small\setlength{\tabcolsep}{7pt}
  \caption{Additional W2G32 results. Bold denotes the better quantized result within each model.}
  \label{tab:w2g32}
  \begin{tabular}{ll rrr}
    \toprule
    Model & Method & PPL$\downarrow$ & Avg.$\uparrow$ & Time (s) \\
    \midrule
    \multirow{3}{*}{Llama-2-7B}
    & BF16 & 5.47 & 0.7038 & -- \\
    & GPTQ & 583.21 & 0.3607 & 266.6 \\
    \rowcolor{gray!12} & JointQ & \textbf{26.33} & \textbf{0.4927} & 1164.1 \\
    \midrule
    \multirow{3}{*}{Llama-3-8B}
    & BF16 & 6.14 & 0.7442 & -- \\
    & GPTQ & 195282.65 & 0.3682 & 292.1 \\
    \rowcolor{gray!12} & JointQ & \textbf{108.77} & \textbf{0.4866} & 1330.4 \\
    \midrule
    \multirow{3}{*}{Qwen3-8B}
    & BF16 & 9.73 & 0.7403 & -- \\
    & GPTQ & 147.46 & 0.4063 & 321.2 \\
    \rowcolor{gray!12} & JointQ & \textbf{45.98} & \textbf{0.5755} & 1971.2 \\
    \midrule
    \multirow{3}{*}{Qwen3-14B}
    & BF16 & 8.65 & 0.7736 & -- \\
    & GPTQ & 23.93 & 0.5094 & 518.4 \\
    \rowcolor{gray!12} & JointQ & \textbf{20.78} & \textbf{0.6200} & 6219.7 \\
    \bottomrule
  \end{tabular}
\end{table*}

JointQ outperforms GPTQ on both metrics for all four models. Its average-accuracy gains are $0.1320$ on Llama-2-7B, $0.1184$ on Llama-3-8B, $0.1692$ on Qwen3-8B, and $0.1106$ on Qwen3-14B. The absolute perplexities likewise show that JointQ mitigates the extreme degradation of GPTQ, particularly on the Llama models. This trend is consistent with the mechanism in Proposition~\ref{prop:separation}, while not constituting a guarantee for end-to-end metrics.

Nevertheless, W2G32 remains substantially below BF16. JointQ's perplexities are $26.33$, $108.77$, $45.98$, and $20.78$, compared with BF16 values of $5.47$, $6.14$, $9.73$, and $8.65$. Its average-accuracy gaps to BF16 are $0.2111$, $0.2576$, $0.1648$, and $0.1536$, respectively. Thus, JointQ markedly improves the 2-bit solution relative to GPTQ, but does not recover practical model quality in this aggressive regime.
\section{Related Work}
\label{sec:related}
\paragraph{Layer-wise solvers.}
Recent advances in post-training quantization (PTQ) for large language models have been shaped largely by layer-wise reconstruction objectives and second-order approximations. OBQ~\citep{frantar2022obc} quantizes weights sequentially using second-order updates, and GPTQ~\citep{frantar2023gptq} extends this line to billion-parameter transformers. CDQuant~\citep{nair2024cdquant} also minimizes reconstruction error through coordinate descent, but it assumes fixed scales throughout the optimization.

\paragraph{Joint optimization of scales and integers.}
The works closest to JointQ treat the scales and the integers of a standard scalar format as joint variables of the layer-wise objective. COMQ~\citep{zhang2025comq} applies coordinate descent to both: it updates the integers one coordinate at a time with the scale held fixed and then updates the scale in closed form, so each integer update is evaluated under a frozen scale, the setting of Corollary~\ref{cor:frozen}. Beacon~\citep{zhang2025beacon} instead substitutes the closed-form optimal scale into the per-channel objective, which reduces the problem to maximizing the absolute value of the cosine of the angle between $wX$ and $qX$, where $q$ is the integer vector. Its search then drops the absolute value: it assigns the integers by a greedy sequential pass followed by cyclic coordinate updates that maximize the signed cosine, and computes the scale in closed form only at the end. In the per-channel case ($T = 1$), JointQ scores its integer moves by this reduced objective rather than by the signed criterion: with $r_0 = w$, the captured energy in Eq.~\eqref{eq:captured} is $\kappa_0(u) = \lVert w \rVert_{\widetilde{H}}^2 \cos^2 \theta_{\widetilde{H}}(u, w^\top)$ whenever $\gamma_0(u) > \varepsilon$ and $w \widetilde{H} w^\top > 0$, where $\lVert w \rVert_{\widetilde{H}}^2 = w \widetilde{H} w^\top$ and $\theta_{\widetilde{H}}(\cdot, \cdot)$ denotes the angle in the $\widetilde{H}$-inner product. This squared cosine ranks candidates as the absolute cosine does, with $\widetilde{H}$ in place of $H$; the two methods also differ in the codebook and in the search procedure. Scoring integer candidates by their best-response scale is therefore not new in this setting, and the closed form in Proposition~\ref{prop:best_scale} and the case $T = 1$ of Theorem~\ref{thm:scale_elim} are known~\citep{zhang2025comq,zhang2025beacon}.

JointQ differs from these methods in two respects. First, it targets group-wise quantization, in which the $T$ scales of a row are coupled through the off-diagonal blocks of $\widetilde{H}$, whereas COMQ and Beacon are formulated with one scale per layer or per output channel. A group-wise extension of COMQ would still evaluate each integer update under frozen scales. A direct group-wise extension of Beacon would either treat the groups independently, discarding the cross-group coupling, or score every candidate move by the reduced objective $F$, which couples all $T$ scales of the row (Theorem~\ref{thm:scale_elim}) and makes each evaluation considerably more expensive. JointQ takes an intermediate route. Its surrogate $\mac{L}_t$ re-optimizes only the scale of the modified group but retains the full $\widetilde{H}$ through the residual $r_t$, so the cross-group coupling enters every move while each move is still scored in $\mac{O}(1)$ time. Its block-coordinate passes revisit every group and are interleaved with joint re-solves of all $T$ scales. Second, JointQ refines the GPTQ solution rather than replacing it: its warm start contains the error-compensated GPTQ solution, and Proposition~\ref{prop:warm_dominance} guarantees that, without regularization, the output never has a larger row-wise calibration error than GPTQ. In the per-channel case, the Clip-Optimize candidate of the warm start is itself a Beacon-type search (Section~\ref{sec:warm_start}). We do not include COMQ or Beacon as baselines; Section~\ref{sec:conclusion} discusses what this leaves open.

\paragraph{Scale selection and continuous parameters.}
PiSO~\citep{amboage2026piso} approaches the same coupling from the scale side. Under round-to-nearest assignment, the integers are piecewise constant in the scale, and PiSO computes the optimal channel-wise scale exactly by sweeping these pieces. It extends the sweep to group-wise quantization by treating the groups independently or sequentially in a single pass, and it interleaves the sweep with GPTQ, which then assigns the integers under the fixed scales. JointQ works in the opposite direction: it changes the integers after GPTQ and re-optimizes the scales for the integers it selects. Two-Stage Grid Optimization~\citep{kim2026twostage} refines group scales while keeping integer assignments fixed. JointQ also uses such fixed-integer refinement: its warm start refits all group scales with the integers fixed, and its local search periodically re-solves them jointly. Our contribution lies not in this refit but in the group-wise coupled search and the GPTQ-anchored design described above. EasyQuant~\citep{wu2020easyquant}, AWQ~\citep{lin2023awq}, OmniQuant~\citep{shao2023omniquant}, AffineQuant~\citep{ma2024affinequant}, and LRQ~\citep{lee2024lrq} optimize scales, clipping thresholds, or related continuous transformations. JointQ differs from these methods in that it does not introduce a new parameterization; instead, it jointly optimizes the quantities already present in the standard GPTQ format.

\paragraph{Rounding under a fixed grid.}
Another closely related direction seeks to improve rounding while holding the quantization grid fixed. Representative methods include AdaRound~\citep{nagel2020adaround}, AdaQuant~\citep{hubara2020adaquant}, BRECQ~\citep{li2021brecq}, SignRound~\citep{cheng2023signround}, GuidedQuant~\citep{kim2025guidedquant}, YAQA~\citep{tseng2025yaqa}, VQRound~\citep{zhou2026vqround}, and regularized successive-rounding~\citep{cha2026rcsr}. Rather than improving rounding under a fixed grid, JointQ addresses the underlying discrete--continuous coupling between integer assignments and group scales.

\paragraph{Other quantization families.}
Finally, several methods pursue alternative quantization families altogether, including QuIP~\citep{chee2023quip}, AQLM~\citep{egiazarian2024aqlm}, GPTVQ~\citep{vanbaalen2024gptvq}, and QTIP~\citep{tseng2024qtip}. By contrast, JointQ targets the standard group-wise uniform scalar quantization format and improves the solver itself within this widely used deployment setting.

\section{Conclusion}
\label{sec:conclusion}
We introduced JointQ, a layer-wise PTQ method that improves quantization within the standard GPTQ deployment format. The central idea is to address a limitation of standard GPTQ-style quantization: continuous group scales and discrete integer assignments are treated separately even though they are tightly coupled. JointQ addresses this mismatch by extending scale-eliminated integer search, previously formulated for a single scale per output channel, to group-wise quantization: each move updates an integer assignment and recomputes the optimal scale of the affected group in closed form while accounting for the other groups through the full Hessian. The search refines the GPTQ solution rather than replacing it, so that its unregularized layer-wise calibration error never exceeds that of GPTQ. The resulting weights retain the standard low-bit representation.

Our analysis provides a mechanism for why this joint treatment can matter when bit-widths are low or groups are large, without claiming a downstream-metric guarantee. Across four models and four quantization configurations, JointQ improves six-task average zero-shot accuracy over GPTQ in 14 of 16 comparisons and perplexity in 12 of 16, with the mean accuracy gain increasing from $0.0044$ in W4G128 to $0.1559$ in W3 per-channel. A staged ablation shows that the unregularized JointQ pipeline reduces layer-wise calibration output error in every case shown, as guaranteed by its construction, while improving perplexity in 10 of 16 model--configuration pairs and average accuracy in 11 of 16. Thus, better layer-local reconstruction alone does not ensure better end-to-end performance. JointQ also improves average accuracy over \gptqstar in all eight per-channel comparisons, although the additional value of JointQ (\gptqstar cand.) in G128 is model dependent. The full stabilization policy further mitigates severe W3 per-channel perplexity degradation and reduces calibration-seed sensitivity in the evaluated low-sample setting. Consequently, the largest gains in the most challenging regimes cannot be attributed to the local search alone. Perplexity remains model dependent, and W2G32 remains far below BF16 despite substantial improvement over GPTQ. Overall, the results support a focused conclusion: strengthening scale--integer optimization helps preserve downstream accuracy when challenging quantization regimes expose the limitations of a frozen grid, while retaining the deployment format practitioners already use.

\paragraph{Limitations.}
Our experiments evaluate JointQ as a complete pipeline. Because the local search starts from a multi-candidate warm start that already refits all group scales and selects candidates per output row, the present results do not separate the contributions of the Clip-Optimize candidate, the global scale refit, the row-wise selection, and the local search. Within the local search, the G128 results also do not separate the per-pass joint scale re-solve from the integer moves. The thresholds $(\tau_{\mathrm{lo}}, \tau_{\mathrm{hi}})$ that select rows for this re-solve were set by a brief preliminary tuning and have not been ablated, although the per-channel results do not depend on them (Section~\ref{sec:local_search}). We have also not studied the effect of the stopping parameters $\omega_{\mathrm{in}}$ and $\omega_{\mathrm{out}}$ or of the iteration cap. Finally, JointQ (\gptqstar cand.) retains the Clip-Optimize candidate and therefore does not measure the effect of a pure \gptqstar warm start. Quantifying these components, for example by evaluating the warm start without the local search or with a single candidate, or the local search without the joint re-solve, is left for future work.

We also do not compare JointQ with COMQ, Beacon, or PiSO~\citep{zhang2025comq,zhang2025beacon,amboage2026piso}. In the per-channel settings, where the gains over GPTQ are largest, the move score of JointQ coincides with Beacon's reduced objective, and the Clip-Optimize candidate is itself a Beacon-type search (Section~\ref{sec:warm_start}). The results therefore do not show whether JointQ improves on such methods, or how much of its per-channel gain over GPTQ is due to the scale-eliminated search rather than to the GPTQ candidate, the row-wise selection, or the regularization. Likewise, we have not compared the group-wise local search with a variant that treats the groups independently, so the benefit of retaining the cross-group coupling has not been measured separately.

The regularization strengths $\lambda$ were selected by a manual preliminary search on Llama-3-8B, Qwen3-8B, and Qwen3-14B using WikiText-2 perplexity and zero-shot accuracy on ARC-c, ARC-e, PIQA, and WinoGrande, which are part of the evaluation data that we report (Appendix~\ref{app:reg_settings}), whereas the baselines use their default settings. The reported results for these three models may therefore be optimistically biased. Llama-2-7B and the HellaSwag and BoolQ tasks, which were not used in the search, provide only a partial check of how the values transfer. A selection protocol based on held-out data, the effect of the exponent $\eta$, which was not tuned, and the choice of the retry threshold and retry strength, including whether a different retry policy would help in W2G32, remain to be studied.

JointQ is also considerably more expensive than GPTQ at quantization time. In the four main settings, its quantization time is $2.63$--$15.71\times$ that of GPTQ (Appendix~\ref{app:pertask}), with the largest ratios on Qwen3-14B, and the ratio depends on the FP64 throughput of the GPU (Section~\ref{sec:experiments}). This cost is incurred once, offline, and the quantized model has the same format and hence the same inference cost as a GPTQ model, but it may matter when many models or configurations must be quantized.
\bibliographystyle{plainnat}
\bibliography{references}
\appendix
\newpage

\section{Proofs}
\label{app:proofs}

This appendix collects the proofs of all theorems, propositions, and corollaries presented in the main text.

\subsection{Global scale elimination -- Theorem~\ref{thm:scale_elim}}
\label{app:scale_elim_proof}

Let $I = I(\B{q}) = \{t : a_t \neq 0\}$ be the set of active groups.
If $I = \emptyset$, then $\widehat{w} = 0$ for any $\B{s}$ and $\phi = w \widetilde{H} w^\top$; every $\B{s}$ functions as a minimizer, yielding $F(\B{q}) = w \widetilde{H} w^\top$.
Now assume $I \neq \emptyset$ and let $A_I(\B{q})$ be the submatrix of active rows.
We establish that $B_I(\B{q}) = A_I \widetilde{H} A_I^\top \succ 0$.
For any $z \in \mab{R}^{1 \times |I|}$ with $z \neq 0$, we compute $z B_I z^\top = (z A_I) \widetilde{H} (z A_I)^\top$.
The vector $z A_I \in \mab{R}^{1 \times p}$ is a linear combination of the active rows of $A(\B{q})$.
Since each active $a_t$ has support on the disjoint set $G_t$ and is nonzero by definition, $z \neq 0$ implies $z A_I \neq 0$.
Since $\widetilde{H} \succ 0$, this leads to $z B_I z^\top > 0$, thereby establishing $B_I \succ 0$.
Inactive groups do not affect the objective; following the canonical convention, we set $s_t^\star = 0$ for $t \notin I$.

Restricting to the active scales $\B{s}_I$ and expanding Eq.~\eqref{eq:phi_def} with $\widehat{w} = \B{s}_I A_I(\B{q})$:
\begin{equation}
  \phi(\B{s}_I, \B{q}) = w\widetilde{H} w^\top - 2 \B{s}_I A_I(\B{q}) \widetilde{H} w^\top + \B{s}_I B_I(\B{q}) \B{s}_I^\top.
\end{equation}
Setting $\nabla_{\B{s}_I} \phi = 0$ yields $\B{s}_I^\star(\B{q}) = w \widetilde{H} A_I(\B{q})^\top B_I(\B{q})^{-1}$.
Completing the square:
\begin{equation}
  \phi(\B{s}_I, \B{q}) = F(\B{q}) + (\B{s}_I - \B{s}_I^\star) B_I(\B{q}) (\B{s}_I - \B{s}_I^\star)^\top.
\end{equation}
Since $B_I(\B{q}) \succ 0$, the second term is non-negative and vanishes only at $\B{s}_I = \B{s}_I^\star$; thus, the active components of the minimizer are uniquely determined. Under the canonical convention $s_t^\star = 0$ for $t \notin I$, this provides the unique canonical representative of the minimizer.

\subsection{Best-response scale -- Proposition~\ref{prop:best_scale}}
\label{app:best_scale_proof}

Eq.~\eqref{eq:one_group_quad} takes the form $\phi(s_t) = \alpha - 2\beta s_t + \gamma s_t^2$ with $\alpha = r_t \widetilde{H} r_t^\top$, $\beta = a_t \widetilde{H} r_t^\top$, and $\gamma = a_t \widetilde{H} a_t^\top$.
When $\gamma > 0$, this is strictly convex, featuring a unique minimizer $s_t^\star = \beta / \gamma$ and a second derivative of $\phi''(s_t) = 2\gamma > 0$, confirming the minimum.
Substituting: $\phi(s_t^\star) = \alpha - \beta^2/\gamma$.
When $\gamma = 0$, since $\widetilde{H} \succeq 0$, we have $0 = a_t \widetilde{H} a_t^\top = \|\widetilde{H}^{1/2} a_t^\top\|_2^2$, hence $\widetilde{H} a_t^\top = 0$, and therefore $\beta = a_t \widetilde{H} r_t^\top = 0$.
The objective $\phi(s_t) = \alpha$ is independent of $s_t$; we choose $s_t^\star = 0$ as the canonical representative.

\subsection{Conditional error decomposition -- Theorem~\ref{thm:decomp}}
\label{app:decomp_proof}

Fix the residual $r_t$ as in Eq.~\eqref{eq:residual_def}.
The group-$t$ conditional error with $x \in \mab{R}^d$ is $\Psi_t(x) = r_t \widetilde{H} r_t^\top - 2 x^\top g_t + x^\top \widetilde{H}_t x$.
Since $\widetilde{H}_t \succ 0$, completing the square gives
\begin{equation}
  \Psi_t(x) = \Psi_t^{\mathrm{cont}} + (x - v_t)^\top \widetilde{H}_t (x - v_t),
\end{equation}
where $v_t = \widetilde{H}_t^{-1} g_t$ and $\Psi_t^{\mathrm{cont}} = r_t \widetilde{H} r_t^\top - g_t^\top \widetilde{H}_t^{-1} g_t$.
Setting $x = s q_t$ and writing $\beta = q_t^\top g_t$, $\gamma = q_t^\top \widetilde{H}_t q_t$:
\begin{align}
  \Psi_t(s q_t) - \Psi_t^{\mathrm{cont}}
  &= (s q_t - v_t)^\top \widetilde{H}_t (s q_t - v_t) \notag \\
  &= \gamma s^2 - 2\beta s + v_t^\top \widetilde{H}_t v_t \notag \\
  &= (v_t^\top \widetilde{H}_t v_t - \beta^2/\gamma) + \gamma(s - \beta/\gamma)^2.
\end{align}
For the first term, note that
\[
  v_t^\top \widetilde{H}_t v_t = g_t^\top \widetilde{H}_t^{-1} g_t = \|v_t\|_{\widetilde{H}_t}^2,
  \qquad
  \beta^2/\gamma = \langle q_t, v_t\rangle_{\widetilde{H}_t}^2 / \|q_t\|_{\widetilde{H}_t}^2,
\]
where the second identity uses $\langle q_t, v_t\rangle_{\widetilde{H}_t} = q_t^\top \widetilde{H}_t v_t = q_t^\top g_t = \beta$. Hence the first term equals the direction gap in projection form.
If $v_t \neq 0$, this is $\|v_t\|_{\widetilde{H}_t}^2 \sin^2\theta$ where $\cos\theta = \langle q_t, v_t\rangle_{\widetilde{H}_t} / (\|q_t\|_{\widetilde{H}_t} \|v_t\|_{\widetilde{H}_t})$; if $v_t = 0$ then $\beta = 0$ and the direction gap is zero.

\subsection{Separation -- Proposition~\ref{prop:separation}}
\label{app:separation_proof}

\paragraph{Initialization.}
With $\widetilde{H} = I_d$, the symmetric min--max scale is $s^{(0)} = M/Q$. Since $M > 2Q$, we have $1/s^{(0)} = Q/M < 1/2$, so the components $w_j = 1$ for $j \ge 1$ round to $0$, while $w_0 / s^{(0)} = Q$ exactly. Thus $q^{(0)} = (Q, 0, \ldots, 0)$. Its best-response scale is $w \cdot q^{(0)} / \lVert q^{(0)}\rVert^2 = MQ/Q^2 = M/Q$, which reproduces $w_0 = M$ exactly, so $E_0 = \lVert w\rVert^2 - M^2 = d - 1$.

\paragraph{Monotone path.}
Define $q^{(k)} = (Q, \underbrace{1, \ldots, 1}_{k}, 0, \ldots, 0)$ for $k = 0, \ldots, d-1$. Every entry lies in $\{0, 1, Q\} \subseteq \mac{C}_b^{\mathrm{sym}}$, and $q^{(k+1)}$ differs from $q^{(k)}$ by the single update $\delta = +1$ at coordinate $k+1$. Write $N_k \coloneqq w \cdot q^{(k)} = MQ + k$ and $D_k \coloneqq \lVert q^{(k)}\rVert^2 = Q^2 + k$. The best-response scale is $s_k = N_k / D_k > 0$, and $E_k = \lVert w\rVert^2 - N_k^2 / D_k$. Then
\begin{equation*}
  E_{k+1} - E_k
  = \frac{N_k^2}{D_k} - \frac{(N_k + 1)^2}{D_k + 1}
  = \frac{N_k^2 - D_k(2N_k + 1)}{D_k (D_k + 1)},
\end{equation*}
and expanding the numerator gives
\begin{equation*}
  N_k^2 - D_k(2N_k + 1)
  = Q^2 (M^2 - 2MQ - 1) - (2Q^2 + 1)k - k^2.
\end{equation*}
The roots of $M^2 - 2MQ - 1$ are $Q \pm \sqrt{Q^2 + 1}$, so $0 < M < Q + \sqrt{Q^2 + 1}$ implies $M^2 - 2MQ - 1 < 0$. The remaining terms are non-positive for $k \ge 0$. Hence every difference is negative, and the path is strictly decreasing.

\paragraph{Endpoint and ratio.}
At $k = d-1$, using $\lVert w\rVert^2 = M^2 + d - 1$,
\begin{equation*}
  (M^2 + d - 1)(Q^2 + d - 1) - (MQ + d - 1)^2 = (M - Q)^2 (d - 1),
\end{equation*}
so $E_{d-1} = (M - Q)^2 (d - 1)/(Q^2 + d - 1)$ and $E_0 / E_{d-1} = (Q^2 + d - 1)/(M - Q)^2$. The admissible range gives $Q < M - Q < \sqrt{Q^2 + 1}$, so $(M - Q)^2 < Q^2 + 1$, which yields the stated lower bound. The admissible range is nonempty because $Q + \sqrt{Q^2 + 1} - 2Q = \sqrt{Q^2 + 1} - Q > 0$.

\paragraph{Reachability by Algorithm~\ref{alg:inner}.}
With $T = 1$ and $\widetilde{H} = I_d$, the residual is $r_0 = w$, so $g_{0,0} = w^\top$, $\alpha_0 = \lVert w\rVert^2$, and $\gamma_0(u) = \lVert u\rVert^2$. By the computation above, the first decrease is $E_0 - E_1 = (1 - M(M - 2Q))/(Q^2 + 1) < 1$ because $M > 2Q$, so the assumption on $\varepsilon$ implies $\varepsilon < 1$. Every nonzero $u \in \mac{C}_b^d$ has $\gamma_0(u) \ge 1 > \varepsilon$, so the guard in Eq.~\eqref{eq:captured} is inactive except at $u = 0$, where $\kappa_0(0) = 0$. Hence $E(u) = \lVert w\rVert^2 - \kappa_0(u)$ for every candidate, and increasing $\kappa_0$ is the same as decreasing $E$.

Fix $0 \le k < d-1$, and suppose that the current iterate is obtained from $q^{(k)}$ by permuting the coordinates $1, \ldots, d-1$. It has the value $Q$ at coordinate $0$, $k$ coordinates equal to $1$, and $d - 1 - k \ge 1$ coordinates equal to $0$, and it has $N = N_k$ and $D = D_k$. Since $\kappa_0 = N^2/D$ for nonzero candidates, the $\pm 1$ updates fall into six classes.
\begin{enumerate}[label=(\alph*), leftmargin=*, itemsep=0pt]
  \item $+1$ at a coordinate with value $0$ gives $(N + 1, D + 1)$. The result is a permutation of $q^{(k+1)}$, so $\kappa_0$ increases by $E_k - E_{k+1} > \varepsilon$.
  \item $+1$ at a coordinate with value $1$, feasible only if $Q \ge 2$, gives $(N + 1, D + 3)$. Its value $\kappa_0$ is smaller than in (a) because $(N + 1)^2 > 0$.
  \item $-1$ at a coordinate with value $0$ gives $(N - 1, D + 1)$. Its value is smaller than in (a) because $N > 0$.
  \item $-1$ at a coordinate with value $1$, possible only if $k \ge 1$, gives a permutation of $q^{(k-1)}$, which has a larger $E$ because the path is strictly decreasing.
  \item $-1$ at coordinate $0$ gives the value $Q - 1$ there. If $k = 0$, the candidate is $(Q - 1, 0, \ldots, 0)$, whose error is $d - 1$ if $Q \ge 2$ and $\lVert w\rVert^2 > d - 1$ if $Q = 1$. If $k \ge 1$, let $h(c) \coloneqq (Mc + k)^2/(c^2 + k)$ for $c \ge 0$. Its derivative has the sign of $k(Mc + k)(M - c)$, which is positive for $0 \le c < M$, so $h(Q - 1) < h(Q)$, the current value of $\kappa_0$. In both cases, the candidate does not increase $\kappa_0$.
  \item $+1$ at coordinate $0$ is infeasible, because $Q = L - 1$ is the upper end of both $\mac{C}_b^{\mathrm{sym}}$ and $\mac{C}_b$.
\end{enumerate}
Every candidate in (a) has the same value. The best $+1$ candidate is therefore in class (a), whichever index the arg max returns, while no $-1$ candidate exceeds the current $\kappa_0$. Algorithm~\ref{alg:inner} thus selects $\delta = +1$ with $\Delta = E_k - E_{k+1} > \varepsilon$. The break condition cannot hold, because no row has $\Delta < \varepsilon$, so the move is applied, and the new iterate is a permutation of $q^{(k+1)}$ that fixes coordinate $0$. Starting from $q^{(0)}$, induction over $k = 0, \ldots, d-2$ shows that the first $d - 1$ iterations reach permutations of $q^{(1)}, \ldots, q^{(d-1)}$; the cap of $10d$ iterations is not reached. Because $\widetilde{H} = I_d$ and $w_j = 1$ for all $j \ge 1$, the error $E$ is invariant under these permutations. Every candidate considered above is feasible in $\mac{C}_b^{\mathrm{sym}}$ if and only if it is feasible in $\mac{C}_b$, because the iterates only take the values $0$, $1$, and $Q$, so the run is the same for both feasibility checks.

\subsection{Descent and termination -- Proposition~\ref{prop:termination}}
\label{app:termination_proof}

\paragraph{Zero-curvature guard.}
By the assumption on $\varepsilon$, $\gamma_t(u) \le \varepsilon$ implies $\gamma_t(u) = 0$, and then $\beta_t(u) = 0$ by the argument in Appendix~\ref{app:best_scale_proof}. Hence $\kappa_t(u)$ in Eq.~\eqref{eq:captured} equals $\beta_t(u)^2/\gamma_t(u)$ if $\gamma_t(u) > 0$ and $0$ otherwise. By Proposition~\ref{prop:best_scale}, $\mac{L}_t(u) = \alpha_t - \kappa_t(u)$ for every $u \in \mac{C}_b^d$, and the scale $\sigma$ returned by Algorithm~\ref{alg:inner} is the best-response scale $s_t^\star(u)$.

\paragraph{(i) Group update.}
Fix a row and a visit to group $t$, and suppress the row index. Before the visit, Eq.~\eqref{eq:one_group_quad} and Proposition~\ref{prop:best_scale} give $\phi = \alpha_t - 2 s_t \beta_t(q_t) + s_t^2 \gamma_t(q_t) \ge \mac{L}_t(q_t)$. Algorithm~\ref{alg:inner} starts from $u = q_t$ with $\kappa = \kappa_t(q_t)$, and Eq.~\eqref{eq:incr} keeps $\beta$, $\gamma$, and $\kappa$ equal to $\beta_t(u)$, $\gamma_t(u)$, and $\kappa_t(u)$ for the current $u$. Each applied move increases $\kappa_t(u)$ by more than $\varepsilon$, so after $k$ moves,
\begin{equation*}
  \mac{L}_t(u) = \alpha_t - \kappa_t(u) \le \mac{L}_t(q_t) - k\varepsilon \le \phi - k\varepsilon,
\end{equation*}
with strict inequality in the first step if $k \ge 1$. If $k \ge 1$, then $u \neq q_t$ because $\kappa_t$ strictly increased, so some row changed its integers and the pair $(u, \sigma)$ of this row is installed. An installed pair changes only group $t$, so the residual $r_t$ is unchanged, and the new objective equals $\mac{L}_t(u)$ because $\sigma = s_t^\star(u)$. If the pair is not installed, then $k = 0$ and $\phi$ is unchanged.

\paragraph{(ii) Joint re-solve.}
By construction, a re-solved scale vector is installed only if it decreases $\phi_i$ by more than $n\varepsilon$, and rows without an installed re-solve keep their objective.

\paragraph{(iii) Monotonicity and termination.}
By (i) and (ii), no step of Algorithm~\ref{alg:jointq} increases any $\phi_i$. Algorithm~\ref{alg:inner} runs at most $10d$ iterations by construction. Let $\varepsilon > 0$ and $\omega_{\mathrm{out}} > 0$, and suppose that the first $N$ passes do not trigger the stopping rule. After each of these passes, at least $\omega_{\mathrm{out}} m$ rows are active, that is, their objectives decreased by more than $n\varepsilon$ during the pass, while no objective increased. Each of these passes therefore decreases $\sum_i \phi_i$ by more than $n\, \omega_{\mathrm{out}}\, m\, \varepsilon$. Since $\widetilde{H} \succeq 0$ implies $\phi_i \ge 0$,
\begin{equation*}
  N\, n\, \omega_{\mathrm{out}}\, m\, \varepsilon \le \sum_i \phi_i^{(0)},
\end{equation*}
and Algorithm~\ref{alg:jointq} stops after at most $N + 1 \le 1 + \sum_i \phi_i^{(0)} / (n\, \omega_{\mathrm{out}}\, m\, \varepsilon)$ passes.

\subsection{Dominance over the GPTQ candidate -- Proposition~\ref{prop:warm_dominance}}
\label{app:warm_dominance_proof}

The first inequality follows from Proposition~\ref{prop:termination}(iii): Algorithm~\ref{alg:jointq} starts from the selected refitted candidate, whose objective is $\phi_i^{(c_i^\star)}$, and never increases $\phi_i$, whatever its stopping rule and thresholds. The second inequality holds because $c_i^\star$ minimizes $\phi_i^{(c)}$ over a pool that contains the GPTQ candidate. The third holds because the refit installs new scales only if they decrease the objective. Finally, the refitted scales are feasible for $\B{q}_i^{\mathrm{G}}$, so $\phi_i^{(\mathrm{G})} \ge \min_{\B{s}} \phi(\B{s}, \B{q}_i^{\mathrm{G}}) = F_i(\B{q}_i^{\mathrm{G}})$ by Theorem~\ref{thm:scale_elim}. When the undamped solution is installed, it equals $\B{s}_I^\star(\B{q}_i^{\mathrm{G}})$, and equality holds.

\subsection{Regularization equivalence}
\label{app:reg_proof}

Define $\widetilde{X} = [X \mid \sqrt{n\lambda} R^{1/2}] \in \mab{R}^{p \times (n+p)}$.
Then $\widetilde{X} \widetilde{X}^\top = H + n\lambda R = \widetilde{H}$, and
\begin{equation}
  \|(W - \widehat{W})\widetilde{X}\|_F^2
  = \|(W - \widehat{W})X\|_F^2
  + n\lambda \|(W - \widehat{W})R^{1/2}\|_F^2,
\end{equation}
which is Eq.~\eqref{eq:reg_obj}.

\subsection{Regularization curvature and sensitivity}
\label{app:reg_corollary}

\begin{corollary}[Curvature and sensitivity bounds]
\label{cor:reg}
Let $R_t=[R]_{G_t,G_t}$. For $\lambda>0$ and any $q_t\in\mac{C}_b^d$,
\begin{equation}
  q_t^\top \widetilde{H}_t q_t
  \ge n\lambda q_t^\top R_t q_t
  = n\lambda\sum_{j\in G_t}\rho_j q_j^2.
  \label{eq:weighted_curvature}
\end{equation}
Consequently, the scale gap is at least $n\lambda(q_t^\top R_tq_t)(s-s_t^\star)^2$.
If $q_t^\top R_tq_t>0$, the scale sensitivity is bounded by
\begin{equation}
  |s_t^\star(r_t)-s_t^\star(r_t')|
  \le
  \frac{\lVert r_t-r_t'\rVert_{\widetilde{H}}}
       {\sqrt{n\lambda q_t^\top R_tq_t}}.
  \label{eq:weighted_sensitivity}
\end{equation}
\end{corollary}
\begin{proof}
Since $\widetilde{H}_t \succeq n\lambda R_t$, Eq.~\eqref{eq:weighted_curvature} follows.
For sensitivity, $s_t^\star(r) = \tilde{q}_t \widetilde{H} r^\top / (\tilde{q}_t \widetilde{H} \tilde{q}_t^\top)$ where $\tilde{q}_t$ is the zero-padded $q_t$.
By Cauchy--Schwarz in the $\widetilde{H}$-inner product: $|\tilde{q}_t \widetilde{H} (r-r')^\top| \le \lVert\tilde{q}_t\rVert_{\widetilde{H}} \lVert r - r'\rVert_{\widetilde{H}}$.
Dividing by $\tilde{q}_t \widetilde{H} \tilde{q}_t^\top = \lVert\tilde{q}_t\rVert_{\widetilde{H}}^2 \ge n\lambda q_t^\top R_tq_t$ gives Eq.~\eqref{eq:weighted_sensitivity}.
\end{proof}
If $q_t$ is supported only on coordinates with $\rho_j=0$, the diagonal regularizer supplies no strictly positive uniform lower bound; this is why the positive-definiteness assumptions in the main text are stated separately.

\section{Extension to asymmetric quantization}
\label{app:asymmetric}

In asymmetric quantization, the stored integers and zero-points satisfy $q_{G_t}^{\mathrm{raw}} \in \{0,\dots,2^b{-}1\}^d$ and $z_t \in \{0,\dots,2^b{-}1\}$.
Define the centered integers $\bar{q}_{G_t} \coloneqq q_{G_t}^{\mathrm{raw}} - z_t \mathbf{1}$ and the zero-padded row embedding $a_t(\bar{q}_t) \in \mab{R}^{1 \times p}$, so that $\widehat{w}_{G_t} = s_t \bar{q}_{G_t}$ and $\widehat{w} = \sum_t s_t a_t$ as in the symmetric case.
The theory below is written in terms of $\bar{q}_t$, while feasibility is enforced through the raw variables $(q_{G_t}^{\mathrm{raw}}, z_t)$.
Because $a_t$ depends on $q_t^{\mathrm{raw}}$ and $z_t$ only through $\bar{q}_t$, the row objective $\phi = (w - \widehat{w}) \widetilde{H} (w - \widehat{w})^\top$ and the residual $r_t = w - \sum_{u \neq t} s_u a_u$ retain their original forms.
The fixed-zero-point format of Section~\ref{sec:group_format} is the special case $z_t \equiv L = 2^{b-1}$, for which $\bar{q}_{G_t} \in \mac{C}_b^d$.

\paragraph{Best-response scale.}
The best-response scale derived in Proposition~\ref{prop:best_scale} carries over verbatim: $s_t^\star = a_t \widetilde{H} r_t^\top / (a_t \widetilde{H} a_t^\top)$ whenever $a_t \widetilde{H} a_t^\top > 0$, regardless of whether $a_t$ was produced by a symmetric or asymmetric codebook. The minimized value $\phi(s_t^\star) = r_t \widetilde{H} r_t^\top - (a_t \widetilde{H} r_t^\top)^2/(a_t \widetilde{H} a_t^\top)$ is likewise unchanged.

\paragraph{Fixed zero points.}
If the zero points are fixed at initialization, the asymmetric problem has the same structure as the problem of Section~\ref{sec:jointq}. The centered integers of group $t$ then range over the set of consecutive integers $\bar{\mac{C}}_t \coloneqq \{-z_t, \ldots, 2^b - 1 - z_t\}$, which depends on the group but stays fixed during the optimization. Algorithms~\ref{alg:jointq}--\ref{alg:warm_start} therefore apply without modification once their feasibility checks use the bounds of $\bar{\mac{C}}_t$ instead of $\mac{C}_b$. Theorem~\ref{thm:scale_elim}, Proposition~\ref{prop:best_scale}, Theorem~\ref{thm:decomp}, Corollary~\ref{cor:frozen}, and Propositions~\ref{prop:termination} and~\ref{prop:warm_dominance} continue to hold with $\mac{C}_b^d$ replaced by $\bar{\mac{C}}_t^d$, because none of them uses the symmetry of the codebook. The zero points can be fixed, for example, by the asymmetric min--max rule
\begin{equation*}
  s_t = \frac{\max_{j \in G_t} w_j - \min_{j \in G_t} w_j}{2^b - 1},
  \qquad
  z_t = \mathrm{clamp}\bigl(\lfloor -\min_{j \in G_t} w_j / s_t \rceil, 0, 2^b - 1\bigr),
\end{equation*}
where $s_t$ is clamped below at $10^{-8}$ so that a constant group does not divide by zero, with $q_j^{\mathrm{raw}} = \mathrm{clamp}(\lfloor w_j / s_t + z_t \rceil, 0, 2^b - 1)$ for the Clip-Optimize candidate, and by the zero points of GPTQ for the GPTQ candidate. Each row keeps the zero points of its selected candidate throughout the local search. We have not evaluated this asymmetric mode in this paper.

\paragraph{Zero-point moves.}
A natural extension also moves the zero points. A move $z_t \leftarrow z_t + \delta_z$ with $\delta_z \in \{-1, +1\}$ and fixed raw integers shifts every centered integer of the group by $-\delta_z$. In the notation of Section~\ref{sec:local_search}, the quantities of Eq.~\eqref{eq:incr} change as
\begin{equation*}
  \beta_t' = \beta_t - \delta_z \mathbf{1}_d^\top g_{i,t},
  \qquad
  \gamma_t' = \gamma_t - 2\delta_z \mathbf{1}_d^\top c_{i,t} + \mathbf{1}_d^\top \widetilde{H}_t \mathbf{1}_d,
\end{equation*}
so such a move can still be scored in $\mac{O}(d)$ time, with $\mathbf{1}_d^\top \widetilde{H}_t \mathbf{1}_d$ precomputed once per group. Because it changes the feasible set $\bar{\mac{C}}_t$ itself, however, it does not fit the fixed-codebook search of Algorithms~\ref{alg:jointq} and~\ref{alg:inner}. Combining zero-point moves with the integer search and evaluating asymmetric quantization are left for future work.

\section{Warm-Start Algorithm and Reproducibility}
\label{app:warm_start}

Algorithm~\ref{alg:warm_start} summarizes the multi-candidate warm start of Section~\ref{sec:warm_start}. Its output is the warm start $(\B{s}^{(0)}, \B{q}^{(0)})$ required by Algorithm~\ref{alg:jointq}.

\begin{algorithm}[t]
  \caption{Multi-candidate warm start for one layer}
  \label{alg:warm_start}
  \small
  \begin{algorithmic}[1]
    \Require $W$, $\widetilde{H}$, groups $\{G_t\}$, codebook $\mac{C}_b$, GPTQ damping $\epsilon$; $\varepsilon$, $\omega_{\mathrm{in}}$
    \For{$t = 0, \dots, T{-}1$} \Comment{Clip-Optimize candidate ($c = 1$)}
      \State $\sigma_i \leftarrow \max(\max_{j \in G_t} |w_{i,j}| / (L-1), 10^{-8})$ for all rows $i$
      \State $u_i \leftarrow \mathrm{clamp}(\lfloor w_{i,G_t}^\top/\sigma_i \rceil, -L, L-1)$ for all rows $i$
      \State $(u_i, \sigma_i)_{i} \leftarrow$ Algorithm~\ref{alg:inner} with $g_{i,t} = \widetilde{H}_t w_{i,G_t}^\top$ and initial $u_i$ \Comment{Eq.~\eqref{eq:group_local}}
      \State $(q_{i,t}^{(1)}, s_{i,t}^{(1)}) \leftarrow (u_i, \sigma_i)$ for all rows $i$
    \EndFor
    \State $(\B{s}^{(2)}, \B{q}^{(2)}) \leftarrow$ GPTQ run on $\widetilde{H}$ with damping $\epsilon$ \Comment{GPTQ candidate ($c = 2$)}
    \For{$c \in \{1, 2\}$ and each row $i$} \Comment{global scale refit}
      \State $\B{s}' \leftarrow$ solution of $B(\B{q}_i^{(c)})\B{s}^\top = A(\B{q}_i^{(c)})\widetilde{H} w_i^\top$, with the damping fallback
      \If{$\phi(\B{s}', \B{q}_i^{(c)}) < \phi(\B{s}_i^{(c)}, \B{q}_i^{(c)}) - n\varepsilon$}
        \State $\B{s}_i^{(c)} \leftarrow \B{s}'$
      \EndIf
      \State $\phi_i^{(c)} \leftarrow \phi(\B{s}_i^{(c)}, \B{q}_i^{(c)})$
    \EndFor
    \For{each row $i$} \Comment{row-wise selection}
      \State $c_i^\star \leftarrow \argmin_c \phi_i^{(c)}$, with $c_i^\star = 1$ on ties;\; $(\B{s}_i^{(0)}, \B{q}_i^{(0)}) \leftarrow (\B{s}_i^{(c_i^\star)}, \B{q}_i^{(c_i^\star)})$
    \EndFor
    \State \Return $(\B{s}^{(0)}, \B{q}^{(0)})$
  \end{algorithmic}
\end{algorithm}

\paragraph{Settings used in all experiments.}
The candidate pool contains exactly the Clip-Optimize candidate and one GPTQ candidate. The GPTQ candidate is vanilla GPTQ (\texttt{actorder=false}, \texttt{mse=false}) for JointQ and JointQ w/o stab., and \gptqstar for JointQ (\gptqstar cand.). Both candidates, the scale refit, the selection, and the local search use the same $\widetilde{H}$; the GPTQ candidate additionally uses the damping $\epsilon = 0.01$, identical to the GPTQ baseline. Candidates are selected per output row by the objective $\phi_i^{(c)}$ after the scale refit. An adaptive retry regenerates the whole warm start under the new $\lambda$. Reported quantization times follow the conventions in Section~\ref{sec:experiments}; in particular, JointQ times include warm-start generation.
\paragraph{Search settings.}
Table~\ref{tab:search_settings} lists the tolerances and thresholds of Algorithms~\ref{alg:jointq}--\ref{alg:warm_start}. They are identical for the local search and for the Clip-Optimize refinement, and they are fixed in all experiments.

\begin{table}[t]
  \centering\small
  \caption{Tolerances and thresholds of Algorithms~\ref{alg:jointq}--\ref{alg:warm_start}, fixed in all experiments.}
  \label{tab:search_settings}
  \begin{tabular}{l l p{0.46\linewidth}}
    \toprule
    Setting & Value & Role \\
    \midrule
    $\varepsilon$ & $10^{-12}$ & Absolute tolerance. Algorithm~\ref{alg:inner} compares it with increases of $\kappa_t$, in the units of $\widetilde{H}$, and uses it as the guard in Eq.~\eqref{eq:captured}; Algorithms~\ref{alg:jointq} and~\ref{alg:warm_start} compare it with decreases of $\phi_i/n$. \\
    Iteration cap & $10d$ & Maximum number of iterations of one call of Algorithm~\ref{alg:inner}. \\
    $\omega_{\mathrm{in}}$ & $0.1$ & Algorithm~\ref{alg:inner} stops when more than this fraction of the rows has $\Delta_i < \varepsilon$. \\
    $\omega_{\mathrm{out}}$ & $0.1$ & Algorithm~\ref{alg:jointq} stops when fewer than this fraction of the rows is active. \\
    $(\tau_{\mathrm{lo}}, \tau_{\mathrm{hi}})$ & $(0.2, 0.6)$ & Selection of the rows for the joint scale re-solve. \\
    $\xi_i$ & $\max(0.01 \max_t [B]_{tt}, 10^{-6})$ & Diagonal damping of the fallback solve. \\
    Scale bound & $65504/L$ & Largest scale magnitude accepted without the damped re-solve. \\
    Ties & $+1$; Clip-Optimize & Preferred move in Algorithm~\ref{alg:inner}; preferred candidate in the row-wise selection. \\
    Precision & float64 & All computations outside the GPTQ run. \\
    \bottomrule
  \end{tabular}
\end{table}

\paragraph{Code release.}
JointQ and all baselines are implemented in the open-source OneComp library~\citep{ichikawa2026onecomp}, whose source code is available at the following URL:
\begin{center}
\url{https://github.com/FujitsuResearch/OneCompression}
\end{center}
\noindent The JointQ quantizer is in \nolinkurl{onecomp/quantizer/jointq} and the GPTQ quantizer in \nolinkurl{onecomp/quantizer/gptq}; the latter is used both for the GPTQ, GPTQ-MSE, and \gptqstar baselines and for the GPTQ candidate of the warm start. The code used to run the experiments is in the \nolinkurl{benchmark} directory of the repository. All results in this paper were obtained with OneComp \href{https://github.com/FujitsuResearch/OneCompression/tree/v1.4.2}{v1.4.2}. Default settings may change in later releases, so this release should be used to reproduce the reported numbers.

\section{Regularization and Retry Settings}
\label{app:reg_settings}

Table~\ref{tab:layerwise_lambda} lists the regularization strengths used by JointQ. This section describes how they, the exponent $\eta$, and the parameters of adaptive retry were chosen.

\begin{table}[t]
  \centering\small\setlength{\tabcolsep}{7pt}
  \caption{Layer-type-specific regularization strengths $\lambda$, shared across model families. All settings use $\eta = 0.5$. In the four main settings, adaptive retry applies to the entries with $\lambda = 0$; it is disabled in W2G32.}
  \label{tab:layerwise_lambda}
  \begin{tabular}{l rrrrr}
    \toprule
    Layer type & W4G128 & W4 per-ch & W3G128 & W3 per-ch & W2G32 \\
    \midrule
    \texttt{q\_proj}    & 0    & 0    & 0.01 & 0.15 & 0.1 \\
    \texttt{k\_proj}    & 0    & 0    & 0.01 & 0.15 & 0.1 \\
    \texttt{v\_proj}    & 0.01 & 0.15 & 0.01 & 0.15 & 0.1 \\
    \texttt{o\_proj}    & 0    & 0    & 0    & 0.15 & 0 \\
    \texttt{gate\_proj} & 0.01 & 0.15 & 0.01 & 0.15 & 0 \\
    \texttt{up\_proj}   & 0.01 & 0.15 & 0.01 & 0.15 & 0 \\
    \texttt{down\_proj} & 0    & 0.01 & 0    & 0.01 & 0 \\
    \bottomrule
  \end{tabular}
\end{table}

\paragraph{Selection of $\lambda$.}
The values in Table~\ref{tab:layerwise_lambda} were chosen by a manual preliminary search, not by an automated procedure. The candidate values were $\{0, 0.01, 0.05, 0.1, 0.15, 0.2,$ $0.3\}$. The search used Llama-3-8B, Qwen3-8B, and Qwen3-14B with the JointQ configuration and the same calibration and evaluation protocol as Section~\ref{sec:experiments}; Llama-2-7B was not used. We first ran each setting with a single $\lambda$ shared by all layer types. For each layer type, we then kept as candidates the values whose unregularized calibration output error was not much larger than with $\lambda = 0$, and chose among them over a few trials. The choice was based on WikiText-2 perplexity and on zero-shot accuracy on ARC-c, ARC-e, PIQA, and WinoGrande; HellaSwag and BoolQ were not used. We preferred configurations that balanced perplexity and accuracy and did not cause a large drop on any of these four tasks. Because these metrics are also reported in Section~\ref{sec:experiments}, $\lambda$ was selected on evaluation data for three of the four models, and their perplexities and accuracies on the four tasks, and hence their six-task averages, may be optimistically biased. Llama-2-7B shares the same values but was not used in the search, so its results show how the selected values transfer to a model outside the search. Relative to JointQ w/o stab., which uses $\lambda = 0$ throughout, JointQ improves the average accuracy of Llama-2-7B in all four main settings and its perplexity in three of them, with a tie in W4G128 (Table~\ref{tab:ablation_reg}). Likewise, HellaSwag and BoolQ were not used for any model, so their scores (Appendix~\ref{app:pertask}) are not directly affected by the selection. On these two tasks, JointQ outperforms GPTQ in 26 of the 32 model--setting--task comparisons, including all 16 comparisons in W3G128 and W3 per-channel; the six losses all occur in 4-bit settings and are at most $0.012$. Neither check is fully held out: the Llama-2-7B results use the same evaluation data as the search, and the HellaSwag and BoolQ results come from values selected partly with WikiText-2 perplexity and the other four tasks. The same values are used without further tuning for JointQ (\gptqstar cand.) and in the calibration-sensitivity study of Appendix~\ref{app:calibration_stability}. The baselines use their default settings and were not tuned on these metrics.

\paragraph{Exponent $\eta$.}
The exponent $\eta = 0.5$ in Eq.~\eqref{eq:diag_regularizer} was fixed before the search above and was not tuned, because it worked adequately in initial runs. We have not evaluated other values.

\paragraph{Adaptive retry.}
The threshold $0.2$ on the relative weight error $\lVert W - \widehat{W}\rVert_F^2 / \lVert W\rVert_F^2$ was set from preliminary W4G128 runs, in which layers at or above this value were outliers among all layers. The retry strength $\lambda = 0.01$ is the smallest nonzero value in the candidate grid. Neither value was tuned further. Because the retry applies only to layers assigned $\lambda = 0$, it can be triggered in W4G128, W4 per-channel, and W3G128, but never in W3 per-channel, where every layer type has $\lambda > 0$. In W2G32, the retry is disabled: with the threshold $0.2$ and retry strength $0.01$, enabling it degraded both perplexity and accuracy in preliminary runs. Other retry parameters might improve the W2G32 results, but we have not ablated them.

\section{Enhanced GPTQ Heuristics}
\label{app:gptqstar}
\label{app:mse_detail}

We denote by \gptqstar an enhanced GPTQ configuration that combines two independent heuristics---activation ordering and MSE-based scale search---on top of the base GPTQ algorithm described in Section~\ref{sec:gptq_background}.

\paragraph{Activation ordering (\texttt{actorder}).}
The base GPTQ processes columns in the natural order $l = 0, 1, \ldots, p-1$.
The \texttt{actorder} heuristic instead processes them in decreasing order of the Hessian diagonal $H_{ll}$, so that the most activation-sensitive columns are quantized first, before the inverse Hessian accumulates error from less important columns.
Concretely, let $\pi$ be a permutation such that $H_{\pi(0),\pi(0)} \ge H_{\pi(1),\pi(1)} \ge \cdots \ge H_{\pi(p-1),\pi(p-1)}$, and let $P$ be the corresponding permutation matrix.
GPTQ is then applied to the permuted system $W' = WP$, $H' = P^\top H P$, and the result is permuted back.
Since the layer objective $\tr((W-\widehat{W})H(W-\widehat{W})^\top)$ is invariant under simultaneous column permutation of $W$, $\widehat{W}$, and $H$, the final reconstruction error is well-defined. With group-wise quantization, the groups are formed contiguously in the permuted order, that is, $G_t = \{\pi(td), \ldots, \pi(td{+}d{-}1)\}$ in the original coordinates.
The cost of the permutation itself is $\mac{O}(mp)$, negligible compared to the $\mac{O}(mp^2)$ GPTQ pass.

\paragraph{MSE-based scale search (\texttt{mse}).}
The base GPTQ sets each group scale by the min--max rule in Eq.~\eqref{eq:scale_minmax} and keeps it fixed thereafter. In per-channel quantization, the scale is computed from $W$ before the column-wise pass begins. In group-wise quantization, it is computed when the pass reaches the first column of the group, from the weights as updated by the error compensation of all previously completed blocks of $B$ columns.
The MSE heuristic replaces the min--max rule, at the same point and on the same weights, with a grid search over $N = 480$ shrinkage factors.
For the symmetric setting used throughout the experiments, the search keeps the centered zero-point fixed and varies only the scale.
For group $t$, each candidate $c = 0, \ldots, N-1$ produces
\begin{equation}
  s_t^{(c)} = \Bigl(1 - \frac{c}{600}\Bigr) \cdot \frac{2 \max_{j \in G_t} |w_j|}{2^b - 1},
  \label{eq:mse_grid}
\end{equation}
and each weight is quantized as $q_j^{(c)} = \mathrm{clamp}(\lfloor w_j / s_t^{(c)} \rceil, -L, L-1)$ and dequantized as $\hat{w}_j^{(c)} = s_t^{(c)} q_j^{(c)}$.
The shrinkage factor $1 - c/600$ thus decreases from $1$, which gives the min--max scale, to $121/600 \approx 0.20$ in steps of $1/600$. The step $1/600$ is the default grid resolution of OneComp, and $N = 0.8 \times 600$ keeps the shrinkage $c/600$ below $0.8$, as in the reference GPTQ implementation.
The selected candidate minimizes a generalized reconstruction metric with exponent $\nu = 2.4$:
\begin{equation}
  c^\star = \argmin_{c \in \{0,\ldots,N-1\}} \sum_{j \in G_t} \bigl|w_j - \hat{w}_j^{(c)}\bigr|^\nu,
  \label{eq:mse_select}
\end{equation}
with ties broken in favor of the smaller $c$, that is, the larger scale.
The exponent $\nu > 2$ penalizes large per-element errors more heavily than the squared loss, which can be advantageous when the weight distribution has outliers within a group.
Note that this heuristic optimizes only the scale; it does not revisit individual integer assignments after selection.

\paragraph{GPTQ-MSE.}
Table~\ref{tab:gptq_mse} gives the model-wise values underlying the summary in Table~\ref{tab:mse_summary}. GPTQ-MSE uses \texttt{mse=true} and \texttt{actorder=false}; JointQ also uses no activation ordering.

\begin{table*}[t]
  \centering\small\setlength{\tabcolsep}{4pt}
  \caption{GPTQ-MSE versus JointQ. Bold denotes the better quality result within each model and setting.}
  \label{tab:gptq_mse}
  \begin{tabular}{ll rr rr}
    \toprule
    & & \multicolumn{2}{c}{GPTQ-MSE} & \multicolumn{2}{c}{JointQ} \\
    \cmidrule(lr){3-4} \cmidrule(lr){5-6}
    Model & Setting & PPL$\downarrow$ & Avg.$\uparrow$ & PPL$\downarrow$ & Avg.$\uparrow$ \\
    \midrule
    \multirow{4}{*}{Llama-2-7B}
    & W4G128 & 5.97 & 0.6913 & \textbf{5.67} & \textbf{0.6914} \\
    & W4 per-ch & 27.21 & 0.5928 & \textbf{5.96} & \textbf{0.6836} \\
    & W3G128 & 7.19 & 0.6568 & \textbf{6.78} & \textbf{0.6610} \\
    & W3 per-ch & 855.39 & 0.4496 & \textbf{7.98} & \textbf{0.6032} \\
    \midrule
    \multirow{4}{*}{Llama-3-8B}
    & W4G128 & 8.84 & 0.7294 & \textbf{6.65} & \textbf{0.7426} \\
    & W4 per-ch & 7.79 & 0.7151 & \textbf{7.41} & \textbf{0.7247} \\
    & W3G128 & 56.76 & 0.6506 & \textbf{8.75} & \textbf{0.7060} \\
    & W3 per-ch & 58.55 & 0.4616 & \textbf{13.09} & \textbf{0.6606} \\
    \midrule
    \multirow{4}{*}{Qwen3-8B}
    & W4G128 & 10.36 & 0.7322 & \textbf{10.17} & \textbf{0.7336} \\
    & W4 per-ch & 12.78 & 0.7084 & \textbf{10.71} & \textbf{0.7240} \\
    & W3G128 & 14.09 & 0.6872 & \textbf{11.50} & \textbf{0.7250} \\
    & W3 per-ch & 51.11 & 0.4950 & \textbf{22.11} & \textbf{0.6686} \\
    \midrule
    \multirow{4}{*}{Qwen3-14B}
    & W4G128 & \textbf{8.79} & \textbf{0.7731} & 8.89 & 0.7699 \\
    & W4 per-ch & \textbf{9.40} & 0.7607 & \textbf{9.40} & \textbf{0.7710} \\
    & W3G128 & \textbf{9.65} & 0.7418 & 9.91 & \textbf{0.7472} \\
    & W3 per-ch & \textbf{17.64} & 0.6725 & 18.92 & \textbf{0.7203} \\
    \bottomrule
  \end{tabular}
\end{table*}

JointQ achieves higher average accuracy in 15 of the 16 comparisons and lower perplexity in 12, with one tie. The accuracy gain is positive for all four models in W3G128 and is largest in the per-channel settings. These results show that stronger one-shot scale selection alone does not reproduce the gains of JointQ.

\paragraph{\gptqstar = \texttt{actorder} + \texttt{mse}.}
Table~\ref{tab:gptqstar} compares \gptqstar with JointQ used in Table~\ref{tab:main_results}. \gptqstar is a useful heuristic configuration but is not uniformly the strongest GPTQ baseline: in some per-channel settings it is worse than GPTQ.

\begin{table*}[t]
  \centering\small\setlength{\tabcolsep}{4pt}
  \caption{\gptqstar versus JointQ. Bold denotes the better quality result within each model and setting.}
  \label{tab:gptqstar}
  \begin{tabular}{ll rrr rrr}
    \toprule
    & & \multicolumn{3}{c}{\gptqstar} & \multicolumn{3}{c}{JointQ} \\
    \cmidrule(lr){3-5} \cmidrule(lr){6-8}
    Model & Setting & PPL$\downarrow$ & Avg.$\uparrow$ & Time (s) & PPL$\downarrow$ & Avg.$\uparrow$ & Time (s) \\
    \midrule
    \multirow{4}{*}{Llama-2-7B}
    & W4G128 & \textbf{5.67} & \textbf{0.6935} & 1301.1 & \textbf{5.67} & 0.6914 & 706.2 \\
    & W4 per-ch & 16.53 & 0.5753 & 362.9 & \textbf{5.96} & \textbf{0.6836} & 1906.5 \\
    & W3G128 & 9.87 & 0.6077 & 1536.4 & \textbf{6.78} & \textbf{0.6610} & 965.9 \\
    & W3 per-ch & 334.30 & 0.4512 & 363.8 & \textbf{7.98} & \textbf{0.6032} & 2161.8 \\
    \midrule
    \multirow{4}{*}{Llama-3-8B}
    & W4G128 & \textbf{6.55} & \textbf{0.7478} & 1454.3 & 6.65 & 0.7426 & 937.0 \\
    & W4 per-ch & 8.19 & 0.7176 & 397.6 & \textbf{7.41} & \textbf{0.7247} & 2564.0 \\
    & W3G128 & \textbf{8.37} & \textbf{0.7110} & 1705.5 & 8.75 & 0.7060 & 1202.8 \\
    & W3 per-ch & 27.55 & 0.6020 & 399.2 & \textbf{13.09} & \textbf{0.6606} & 2736.9 \\
    \midrule
    \multirow{4}{*}{Qwen3-8B}
    & W4G128 & \textbf{9.93} & \textbf{0.7355} & 1546.3 & 10.17 & 0.7336 & 2019.8 \\
    & W4 per-ch & 11.18 & 0.7234 & 414.3 & \textbf{10.71} & \textbf{0.7240} & 1631.7 \\
    & W3G128 & \textbf{11.42} & \textbf{0.7255} & 1824.0 & 11.50 & 0.7250 & 1997.8 \\
    & W3 per-ch & 43.42 & 0.5338 & 417.3 & \textbf{22.11} & \textbf{0.6686} & 2236.2 \\
    \midrule
    \multirow{4}{*}{Qwen3-14B}
    & W4G128 & 8.92 & \textbf{0.7705} & 2610.6 & \textbf{8.89} & 0.7699 & 7435.3 \\
    & W4 per-ch & \textbf{9.26} & 0.7626 & 698.2 & 9.40 & \textbf{0.7710} & 3923.5 \\
    & W3G128 & \textbf{9.52} & \textbf{0.7611} & 3104.6 & 9.91 & 0.7472 & 6937.9 \\
    & W3 per-ch & \textbf{15.68} & 0.6942 & 694.0 & 18.92 & \textbf{0.7203} & 5184.3 \\
    \bottomrule
  \end{tabular}
\end{table*}

JointQ achieves higher average accuracy than \gptqstar in 9 of the 16 comparisons. The split by grouping granularity is more informative: JointQ wins only 1 of 8 G128 comparisons but all 8 per-channel comparisons. The W3 per-channel gains are positive for every model, ranging from $0.0261$ on Qwen3-14B to $0.1520$ on Llama-2-7B. Thus, the two heuristics are often effective in G128, while joint scale--integer optimization is most consistently beneficial when a single scale is shared across an entire channel.

Perplexity is again less uniform: JointQ wins 8 comparisons, ties 1, and loses 7. In particular, Qwen3-14B favors \gptqstar in three of the four perplexity comparisons despite favoring JointQ in both per-channel accuracy comparisons. \gptqstar is also not uniformly stronger than GPTQ; for example, its Qwen3-8B W3 per-channel perplexity is $43.42$, compared with $19.55$ for GPTQ in Table~\ref{tab:main_results}. These results reinforce the distinction between language-modeling perplexity and downstream task accuracy.

\paragraph{JointQ (\gptqstar cand.).}
For the G128 settings, we further evaluate JointQ (\gptqstar cand.), which replaces the vanilla GPTQ candidate in the warm start of Section~\ref{sec:warm_start} with a \gptqstar candidate. When activation ordering is enabled, the columns of the layer are permuted once, before any candidate is generated, and the groups are formed contiguously in the permuted order; the Clip-Optimize candidate, the \gptqstar candidate, the row-wise selection, and the local search all operate on this common grouping. The Clip-Optimize candidate remains in the pool, so this configuration is not a pure \gptqstar warm start. The layer-specific regularization and adaptive retry are retained. The results are mixed: JointQ (\gptqstar cand.) improves perplexity in 4 of 8 cases, ties in 1, and improves average accuracy in 5 of 8. Its total quantization time is $1.32$--$1.45\times$ that of \gptqstar. Hence, in G128, the additional value of JointQ (\gptqstar cand.) remains model dependent.

\begin{table*}[t]
  \centering\small\setlength{\tabcolsep}{4pt}
  \caption{\gptqstar versus JointQ (\gptqstar cand.) in G128. Bold denotes the better quality result within each model and setting.}
  \label{tab:gptqstar_same_init}
  \begin{tabular}{ll rrr rrr}
    \toprule
    & & \multicolumn{3}{c}{\gptqstar} & \multicolumn{3}{c}{JointQ (\gptqstar cand.)} \\
    \cmidrule(lr){3-5} \cmidrule(lr){6-8}
    Model & Setting & PPL$\downarrow$ & Avg.$\uparrow$ & Time (s) & PPL$\downarrow$ & Avg.$\uparrow$ & Time (s) \\
    \midrule
    \multirow{2}{*}{Llama-2-7B}
    & W4G128 & 5.67 & 0.6935 & 1301.1 & \textbf{5.66} & \textbf{0.6960} & 1805.4 \\
    & W3G128 & 9.87 & 0.6077 & 1536.4 & \textbf{6.56} & \textbf{0.6639} & 2226.3 \\
    \midrule
    \multirow{2}{*}{Llama-3-8B}
    & W4G128 & \textbf{6.55} & \textbf{0.7478} & 1454.3 & \textbf{6.55} & 0.7439 & 1913.8 \\
    & W3G128 & \textbf{8.37} & 0.7110 & 1705.5 & 8.41 & \textbf{0.7191} & 2341.4 \\
    \midrule
    \multirow{2}{*}{Qwen3-8B}
    & W4G128 & 9.93 & 0.7355 & 1546.3 & \textbf{9.88} & \textbf{0.7382} & 2089.2 \\
    & W3G128 & \textbf{11.42} & \textbf{0.7255} & 1824.0 & 11.58 & 0.7193 & 2549.3 \\
    \midrule
    \multirow{2}{*}{Qwen3-14B}
    & W4G128 & \textbf{8.92} & 0.7705 & 2610.6 & 8.93 & \textbf{0.7741} & 3747.3 \\
    & W3G128 & 9.52 & \textbf{0.7611} & 3104.6 & \textbf{9.48} & 0.7561 & 4499.6 \\
    \bottomrule
  \end{tabular}
\end{table*}

\section{Layer-wise Output Error}
\label{app:layer_error}

\begin{figure*}[t]
  \centering
  \includegraphics[width=\textwidth]{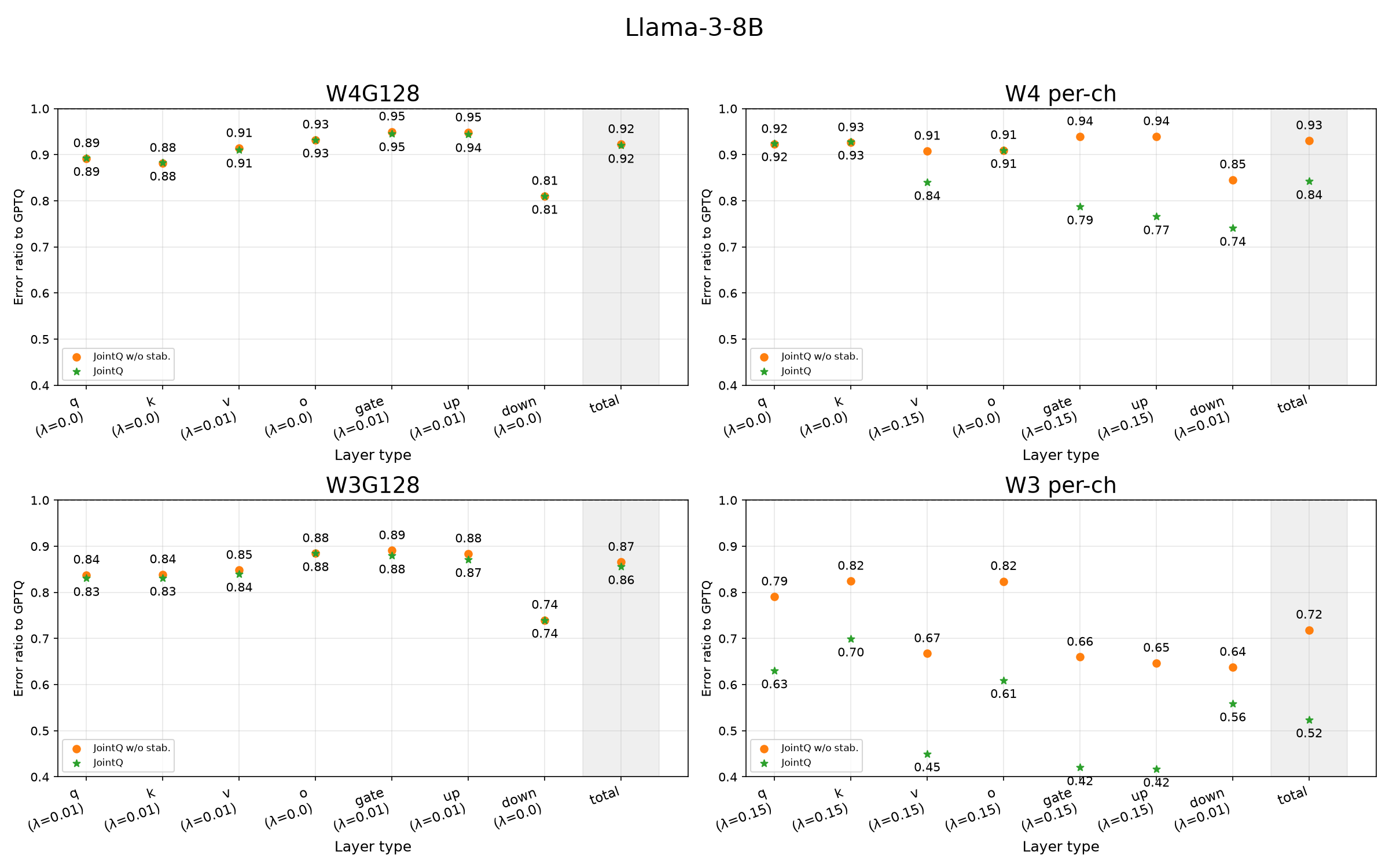}
  \caption{Layer-type-wise unregularized output-error ratios for Llama-3-8B.}
  \label{fig:layer_error_llama3}
\end{figure*}

\begin{figure*}[t]
  \centering
  \includegraphics[width=\textwidth]{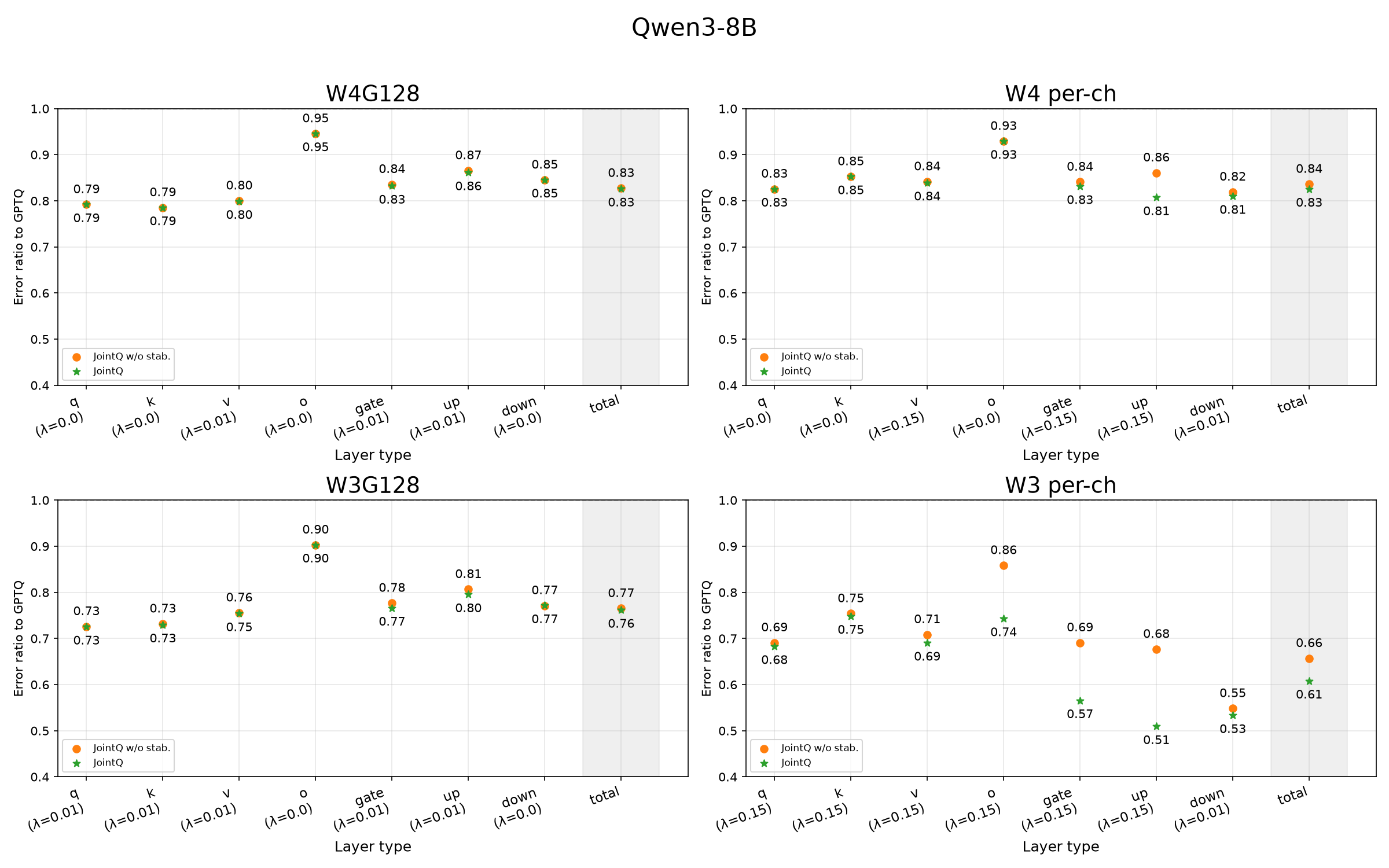}
  \caption{Layer-type-wise unregularized output-error ratios for Qwen3-8B.}
  \label{fig:layer_error_qwen8}
\end{figure*}

\begin{figure*}[t]
  \centering
  \includegraphics[width=\textwidth]{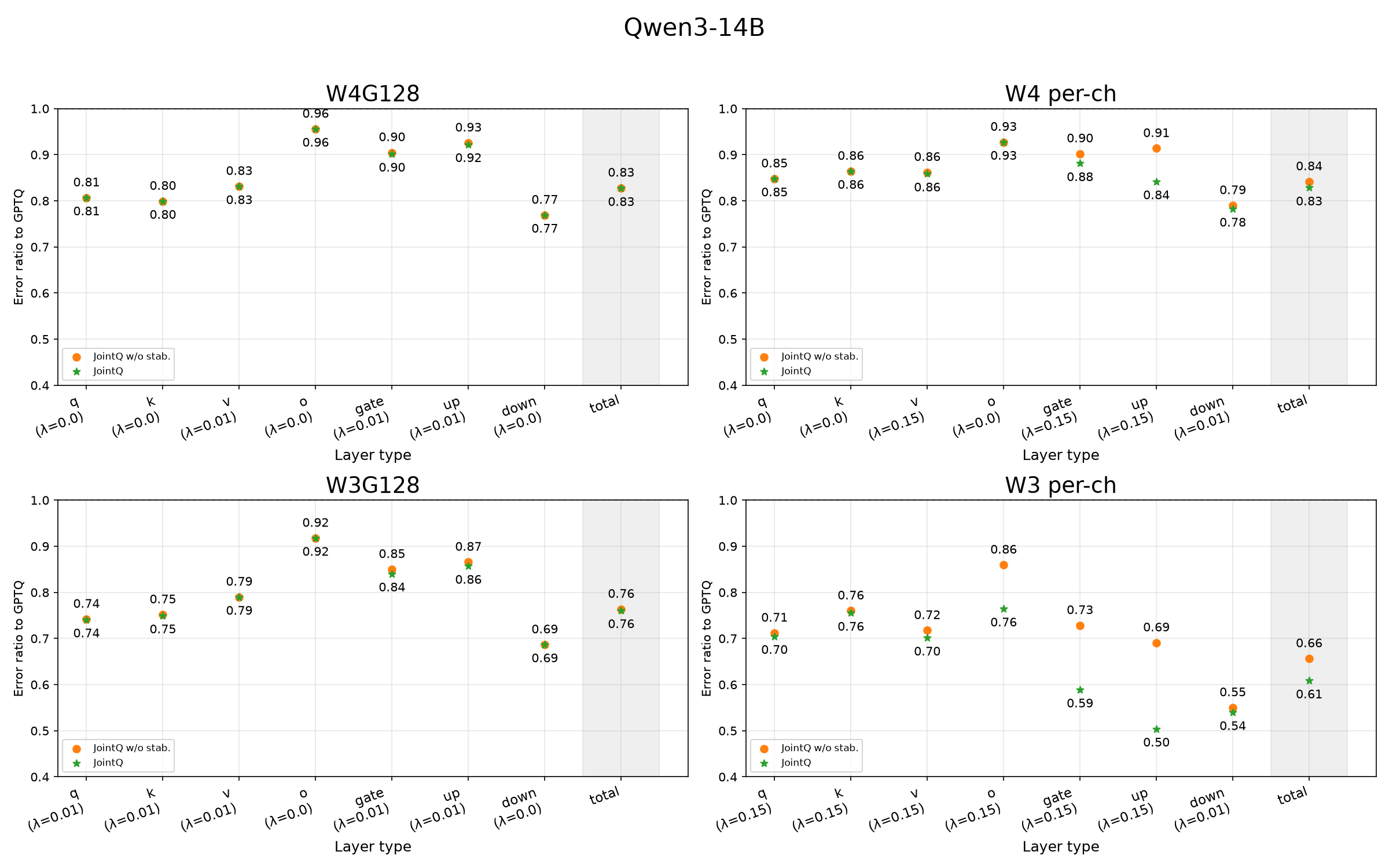}
  \caption{Layer-type-wise unregularized output-error ratios for Qwen3-14B.}
  \label{fig:layer_error_qwen14}
\end{figure*}

Figures~\ref{fig:layer_error_llama3}--\ref{fig:layer_error_qwen14} complement the Llama-2-7B results in Figure~\ref{fig:layer_error_llama2}. At each layer, $X$ is the BF16-model calibration activation used by all methods and to form $H=XX^\top$ for quantization. For each layer type, we sum the unregularized calibration output error $\lVert(W-\widehat W)X\rVert_F^2$ over all transformer blocks and divide by the corresponding GPTQ sum. The ``total'' entry sums all seven layer types before taking the ratio; a ratio below one indicates a lower error than GPTQ.

\section{Calibration-set Stability}
\label{app:calibration_stability}

This section provides the full results for the calibration-sensitivity study in Section~\ref{subsec:calibration_sensitivity}. We evaluate sensitivity to calibration-set sampling on Llama-3-8B in W3G128 and W4G128. Each calibration example is a sequence of length 2048; the sequence counts below are therefore distinct from the calibration-token dimension $n$ used in the theory. For each sequence count, seeds 0--19 share paired calibration data across methods. We report three complementary comparisons: JointQ against \gptqstar, the contribution of JointQ's stabilization policy, and JointQ (\gptqstar cand.) versus \gptqstar.

\paragraph{Recommended systems in W3G128.}
Table~\ref{tab:calibration_system} compares JointQ with \gptqstar. This comparison directly measures the sensitivity of the recommended systems.

\begin{table*}[t]
  \centering\scriptsize\setlength{\tabcolsep}{3pt}
  \caption{System-level calibration sensitivity in W3G128. Values are mean $\pm$ sample standard deviation over 20 paired seeds. Avg. is the six-task mean accuracy. Bold denotes lower mean PPL or higher mean accuracy.}
  \label{tab:calibration_system}
  \begin{tabular}{r rr rr}
    \toprule
    & \multicolumn{2}{c}{JointQ} & \multicolumn{2}{c}{\gptqstar} \\
    \cmidrule(lr){2-3} \cmidrule(lr){4-5}
    Sequences & PPL & Avg. & PPL & Avg. \\
    \midrule
    8   & $\bm{11.0952 \pm 0.2984}$ & $0.6303 \pm 0.0131$ & $11.3662 \pm 1.6689$ & $\bm{0.6322 \pm 0.0297}$ \\
    16  & $\bm{9.9750 \pm 0.2107}$  & $\bm{0.6629 \pm 0.0093}$ & $10.0652 \pm 0.7537$ & $0.6561 \pm 0.0257$ \\
    32  & $\bm{9.3967 \pm 0.1010}$  & $0.6758 \pm 0.0042$ & $9.8167 \pm 0.7994$  & $\bm{0.6772 \pm 0.0173}$ \\
    64  & $9.1449 \pm 0.0786$  & $\bm{0.6828 \pm 0.0054}$ & $\bm{9.0630 \pm 0.3937}$  & $0.6826 \pm 0.0201$ \\
    128 & $8.9921 \pm 0.0838$  & $0.6891 \pm 0.0058$ & $\bm{8.8243 \pm 0.4315}$  & $\bm{0.6915 \pm 0.0184}$ \\
    \bottomrule
  \end{tabular}
\end{table*}

JointQ has lower PPL variance at all five sequence counts. It also has lower mean PPL for 8--32 sequences, whereas \gptqstar is slightly better for 64 and 128 sequences. This comparison supports the low-sample robustness of JointQ.

\paragraph{Stabilization-policy ablation.}
JointQ in Table~\ref{tab:calibration_policy} uses the layer-specific regularization and adaptive retry from Table~\ref{tab:layerwise_lambda}; JointQ w/o stab. disables both. PPL values are mean $\pm$ sample standard deviation, and accuracy differences are reported in percentage points.

\begin{table*}[t]
  \centering\scriptsize\setlength{\tabcolsep}{3pt}
  \caption{Effect of the JointQ stabilization policy. $\Delta$ is JointQ minus JointQ w/o stab.; lower $\Delta$ PPL and higher $\Delta$ Avg. favor JointQ. Wins are counted over 20 paired seeds. Bold denotes the lower mean PPL.}
  \label{tab:calibration_policy}
  \begin{tabular}{rr rrrrrr}
    \toprule
    Bits & Sequences & JointQ PPL & JointQ w/o stab. PPL & $\Delta$ PPL & PPL wins & $\Delta$ Avg. (pt.) & Avg. wins \\
    \midrule
    \multirow{5}{*}{4} & 8   & $\bm{6.9823 \pm 0.2164}$ & $7.0707 \pm 0.2338$ & $-0.0883$ & 20/20 & $+0.371$ & 16/20 \\
      & 16  & $\bm{6.7973 \pm 0.0368}$ & $6.8603 \pm 0.0953$ & $-0.0630$ & 20/20 & $+0.123$ & 12/20 \\
      & 32  & $\bm{6.7211 \pm 0.0199}$ & $6.7575 \pm 0.0304$ & $-0.0365$ & 20/20 & $+0.183$ & 16/20 \\
      & 64  & $\bm{6.6843 \pm 0.0142}$ & $6.7135 \pm 0.0151$ & $-0.0292$ & 19/20 & $+0.260$ & 13/20 \\
      & 128 & $\bm{6.6657 \pm 0.0137}$ & $6.6892 \pm 0.0141$ & $-0.0235$ & 20/20 & $+0.176$ & 14/20 \\
    \midrule
    \multirow{5}{*}{3} & 8   & $\bm{11.0952 \pm 0.2984}$ & $12.7257 \pm 1.0680$ & $-1.6305$ & 20/20 & $+3.185$ & 20/20 \\
      & 16  & $\bm{9.9750 \pm 0.2107}$ & $10.5154 \pm 0.3841$ & $-0.5404$ & 20/20 & $+1.743$ & 19/20 \\
      & 32  & $\bm{9.3967 \pm 0.1010}$ & $9.7241 \pm 0.1355$ & $-0.3274$ & 20/20 & $+0.669$ & 15/20 \\
      & 64  & $\bm{9.1449 \pm 0.0786}$ & $9.3636 \pm 0.0827$ & $-0.2187$ & 20/20 & $+0.388$ & 13/20 \\
      & 128 & $\bm{8.9921 \pm 0.0838}$ & $9.1804 \pm 0.0692$ & $-0.1883$ & 20/20 & $+0.205$ & 12/20 \\
    \bottomrule
  \end{tabular}
\end{table*}

The full policy improves PPL in every W3G128 run and all but one W4G128 run. The effect is strongest with eight W3G128 sequences, where it reduces mean PPL by $1.6305$, reduces PPL standard deviation from $1.0680$ to $0.2984$, and improves mean accuracy by $3.185$ points. The diminishing differences as the sequence count increases indicate that stabilization is most valuable when calibration data are scarce.

\paragraph{JointQ (\gptqstar cand.) versus \gptqstar.}
Table~\ref{tab:calibration_aligned} compares JointQ (\gptqstar cand.) and \gptqstar. Both enable MSE-based scale search and activation ordering, while JointQ (\gptqstar cand.) additionally retains its Clip-Optimize candidate and its stabilization policy. Differences are JointQ (\gptqstar cand.) minus \gptqstar, and the confidence intervals are paired 95\% $t$ intervals over 20 seeds.

\begin{table*}[t]
  \centering\scriptsize\setlength{\tabcolsep}{3pt}
  \caption{JointQ (\gptqstar cand.) versus \gptqstar. Negative $\Delta$ PPL and positive $\Delta$ Avg. favor JointQ (\gptqstar cand.). Wins are counted over 20 paired seeds.}
  \label{tab:calibration_aligned}
  \begin{tabular}{rr rrrr}
    \toprule
    Bits & Sequences & $\Delta$ PPL (95\% CI) & PPL wins & $\Delta$ Avg. points (95\% CI) & Avg. wins \\
    \midrule
    \multirow{5}{*}{4} & 8   & $-0.0387$ [$-0.0552,-0.0221$] & 18/20 & $+0.140$ [$-0.070,+0.349$] & 12/20 \\
      & 16  & $-0.0327$ [$-0.0399,-0.0255$] & 20/20 & $+0.096$ [$-0.064,+0.256$] & 13/20 \\
      & 32  & $-0.0207$ [$-0.0271,-0.0142$] & 18/20 & $+0.049$ [$-0.145,+0.242$] & 10/20 \\
      & 64  & $-0.0202$ [$-0.0273,-0.0131$] & 19/20 & $-0.046$ [$-0.194,+0.102$] & 9/20 \\
      & 128 & $-0.0211$ [$-0.0261,-0.0162$] & 19/20 & $+0.071$ [$-0.062,+0.205$] & 14/20 \\
    \midrule
    \multirow{5}{*}{3} & 8   & $+1.9338$ [$+0.4762,+3.3914$] & 2/20 & $+2.387$ [$+1.433,+3.340$] & 20/20 \\
      & 16  & $+0.6937$ [$+0.1803,+1.2071$] & 6/20 & $+2.202$ [$+1.175,+3.230$] & 19/20 \\
      & 32  & $+1.0020$ [$+0.5276,+1.4763$] & 2/20 & $+1.632$ [$+0.695,+2.569$] & 16/20 \\
      & 64  & $+1.1658$ [$+0.6928,+1.6388$] & 1/20 & $+2.057$ [$+0.979,+3.135$] & 17/20 \\
      & 128 & $+1.4316$ [$+1.1041,+1.7591$] & 0/20 & $+1.358$ [$+0.568,+2.148$] & 17/20 \\
    \bottomrule
  \end{tabular}
\end{table*}

In W4G128, JointQ (\gptqstar cand.) reduces PPL at every sequence count and wins 94 of 100 paired runs; all five PPL confidence intervals exclude zero, whereas the accuracy intervals include zero. In W3G128, both the PPL and accuracy differences are positive at every sequence count, showing that JointQ (\gptqstar cand.) trades higher PPL for higher downstream accuracy. Overall, these results characterize calibration sensitivity on Llama-3-8B in G128 with sequence length 2048.

\section{Detailed Per-Task Results}
\label{app:pertask}

Tables~\ref{tab:llama2_detail}--\ref{tab:qwen3_14b_detail} report the individual zero-shot task scores underlying the averages in Table~\ref{tab:main_results}. ARC-c, ARC-e, PIQA, and HellaSwag use normalized accuracy; WinoGrande and BoolQ use raw accuracy. Because the tasks have different chance levels and sampling uncertainty, small cell-wise differences should not be overinterpreted.

\begin{table*}[ht]
  \centering\scriptsize\setlength{\tabcolsep}{2.5pt}
  \caption{Detailed results for Llama-2-7B.}
  \label{tab:llama2_detail}
  \begin{tabular}{ll r rrrrrr r r}
    \toprule
    Method & Setting & PPL$\downarrow$ & ARC-c & ARC-e & PIQA & WG & HS & BQ & Avg.$\uparrow$ & Time (s) \\
    \midrule
    BF16 & -- & 5.47 & 0.4480 & 0.7378 & 0.7845 & 0.6961 & 0.7617 & 0.7945 & 0.7038 & -- \\
    \midrule
    GPTQ & W4G128 & 5.89 & 0.4437 & 0.7176 & 0.7824 & 0.6867 & 0.7551 & 0.7654 & 0.6918 & 268.1 \\
    \rowcolor{gray!12} JointQ & W4G128 & 5.67 & 0.4352 & 0.7176 & 0.7797 & 0.6961 & 0.7510 & 0.7688 & 0.6914 & 706.2 \\
    \midrule
    GPTQ & W4 per-ch & 6.17 & 0.4369 & 0.6982 & 0.7824 & 0.6827 & 0.7401 & 0.7492 & 0.6816 & 260.3 \\
    \rowcolor{gray!12} JointQ & W4 per-ch & 5.96 & 0.4480 & 0.7113 & 0.7791 & 0.6922 & 0.7336 & 0.7373 & 0.6836 & 1906.5 \\
    \midrule
    GPTQ & W3G128 & 6.80 & 0.3857 & 0.6170 & 0.7563 & 0.6496 & 0.6906 & 0.7162 & 0.6359 & 263.1 \\
    \rowcolor{gray!12} JointQ & W3G128 & 6.78 & 0.3959 & 0.6894 & 0.7693 & 0.6882 & 0.6926 & 0.7306 & 0.6610 & 965.9 \\
    \midrule
    GPTQ & W3 per-ch & 27.74 & 0.2782 & 0.4125 & 0.6572 & 0.5343 & 0.5154 & 0.6144 & 0.5020 & 257.6 \\
    \rowcolor{gray!12} JointQ & W3 per-ch & 7.98 & 0.3601 & 0.5934 & 0.7405 & 0.6267 & 0.6236 & 0.6749 & 0.6032 & 2161.8 \\
    \bottomrule
  \end{tabular}
\end{table*}

\begin{table*}[ht]
  \centering\scriptsize\setlength{\tabcolsep}{2.5pt}
  \caption{Detailed results for Llama-3-8B.}
  \label{tab:llama3_detail}
  \begin{tabular}{ll r rrrrrr r r}
    \toprule
    Method & Setting & PPL$\downarrow$ & ARC-c & ARC-e & PIQA & WG & HS & BQ & Avg.$\uparrow$ & Time (s) \\
    \midrule
    BF16 & -- & 6.14 & 0.5384 & 0.7757 & 0.8058 & 0.7332 & 0.7921 & 0.8199 & 0.7442 & -- \\
    \midrule
    GPTQ & W4G128 & 14.09 & 0.5060 & 0.7559 & 0.7824 & 0.7167 & 0.7810 & 0.8070 & 0.7248 & 277.9 \\
    \rowcolor{gray!12} JointQ & W4G128 & 6.65 & 0.5401 & 0.7828 & 0.7965 & 0.7293 & 0.7843 & 0.8226 & 0.7426 & 937.0 \\
    \midrule
    GPTQ & W4 per-ch & 399.02 & 0.3268 & 0.5206 & 0.6763 & 0.6464 & 0.6916 & 0.7144 & 0.5960 & 273.1 \\
    \rowcolor{gray!12} JointQ & W4 per-ch & 7.41 & 0.4940 & 0.7580 & 0.7829 & 0.7301 & 0.7739 & 0.8092 & 0.7247 & 2564.0 \\
    \midrule
    GPTQ & W3G128 & 57.69 & 0.2628 & 0.4482 & 0.6153 & 0.5943 & 0.4888 & 0.6122 & 0.5036 & 274.8 \\
    \rowcolor{gray!12} JointQ & W3G128 & 8.75 & 0.4693 & 0.7168 & 0.7742 & 0.7395 & 0.7424 & 0.7936 & 0.7060 & 1202.8 \\
    \midrule
    GPTQ & W3 per-ch & 3381.46 & 0.2312 & 0.2753 & 0.5305 & 0.5107 & 0.2752 & 0.4165 & 0.3732 & 269.4 \\
    \rowcolor{gray!12} JointQ & W3 per-ch & 13.09 & 0.4010 & 0.6334 & 0.7388 & 0.7056 & 0.7065 & 0.7783 & 0.6606 & 2736.9 \\
    \bottomrule
  \end{tabular}
\end{table*}

\begin{table*}[ht]
  \centering\scriptsize\setlength{\tabcolsep}{2.5pt}
  \caption{Detailed results for Qwen3-8B.}
  \label{tab:qwen3_8b_detail}
  \begin{tabular}{ll r rrrrrr r r}
    \toprule
    Method & Setting & PPL$\downarrow$ & ARC-c & ARC-e & PIQA & WG & HS & BQ & Avg.$\uparrow$ & Time (s) \\
    \midrule
    BF16 & -- & 9.73 & 0.5640 & 0.8093 & 0.7758 & 0.6772 & 0.7489 & 0.8664 & 0.7403 & -- \\
    \midrule
    GPTQ & W4G128 & 10.33 & 0.5614 & 0.7879 & 0.7639 & 0.6693 & 0.7404 & 0.8709 & 0.7323 & 321.0 \\
    \rowcolor{gray!12} JointQ & W4G128 & 10.17 & 0.5572 & 0.7950 & 0.7628 & 0.6756 & 0.7393 & 0.8719 & 0.7336 & 2019.8 \\
    \midrule
    GPTQ & W4 per-ch & 11.04 & 0.5077 & 0.7441 & 0.7557 & 0.6835 & 0.7246 & 0.8529 & 0.7114 & 315.2 \\
    \rowcolor{gray!12} JointQ & W4 per-ch & 10.71 & 0.5367 & 0.7740 & 0.7688 & 0.6859 & 0.7323 & 0.8465 & 0.7240 & 1631.7 \\
    \midrule
    GPTQ & W3G128 & 11.75 & 0.4881 & 0.7193 & 0.7584 & 0.6527 & 0.6914 & 0.8413 & 0.6919 & 318.5 \\
    \rowcolor{gray!12} JointQ & W3G128 & 11.50 & 0.5375 & 0.7950 & 0.7666 & 0.6882 & 0.6968 & 0.8661 & 0.7250 & 1997.8 \\
    \midrule
    GPTQ & W3 per-ch & 19.55 & 0.3217 & 0.4735 & 0.6850 & 0.5391 & 0.5814 & 0.6691 & 0.5449 & 310.9 \\
    \rowcolor{gray!12} JointQ & W3 per-ch & 22.11 & 0.4582 & 0.7104 & 0.7345 & 0.6448 & 0.6366 & 0.8269 & 0.6686 & 2236.2 \\
    \bottomrule
  \end{tabular}
\end{table*}

\begin{table*}[ht]
  \centering\scriptsize\setlength{\tabcolsep}{2.5pt}
  \caption{Detailed results for Qwen3-14B.}
  \label{tab:qwen3_14b_detail}
  \begin{tabular}{ll r rrrrrr r r}
    \toprule
    Method & Setting & PPL$\downarrow$ & ARC-c & ARC-e & PIQA & WG & HS & BQ & Avg.$\uparrow$ & Time (s) \\
    \midrule
    BF16 & -- & 8.65 & 0.6041 & 0.8279 & 0.7976 & 0.7309 & 0.7877 & 0.8936 & 0.7736 & -- \\
    \midrule
    GPTQ & W4G128 & 8.85 & 0.6015 & 0.8224 & 0.8003 & 0.7301 & 0.7813 & 0.8896 & 0.7709 & 473.2 \\
    \rowcolor{gray!12} JointQ & W4G128 & 8.89 & 0.5998 & 0.8245 & 0.7938 & 0.7269 & 0.7806 & 0.8939 & 0.7699 & 7435.3 \\
    \midrule
    GPTQ & W4 per-ch & 9.11 & 0.5887 & 0.8001 & 0.7840 & 0.7040 & 0.7722 & 0.8657 & 0.7525 & 463.9 \\
    \rowcolor{gray!12} JointQ & W4 per-ch & 9.40 & 0.5956 & 0.8266 & 0.8020 & 0.7324 & 0.7767 & 0.8930 & 0.7710 & 3923.5 \\
    \midrule
    GPTQ & W3G128 & 10.07 & 0.5478 & 0.7862 & 0.7889 & 0.6914 & 0.7489 & 0.8508 & 0.7356 & 461.7 \\
    \rowcolor{gray!12} JointQ & W3G128 & 9.91 & 0.5486 & 0.7959 & 0.7873 & 0.7214 & 0.7526 & 0.8774 & 0.7472 & 6937.9 \\
    \midrule
    GPTQ & W3 per-ch & 13.84 & 0.3720 & 0.5669 & 0.7291 & 0.5746 & 0.6839 & 0.7266 & 0.6089 & 455.2 \\
    \rowcolor{gray!12} JointQ & W3 per-ch & 18.92 & 0.5162 & 0.7504 & 0.7709 & 0.7080 & 0.7141 & 0.8621 & 0.7203 & 5184.3 \\
    \bottomrule
  \end{tabular}
\end{table*}
\end{document}